\documentclass[11pt]{article}

\usepackage[final]{acl}

\usepackage{times}
\usepackage{latexsym}

\usepackage[T1]{fontenc}
\usepackage[utf8]{inputenc}

\usepackage{microtype}

\usepackage{inconsolata}

\usepackage{graphicx}

\usepackage{booktabs}
\usepackage{multirow}
\usepackage{xcolor}
\usepackage{colortbl}
\usepackage{array}

\renewcommand{\thefootnote}{\fnsymbol{footnote}}

\title{EVE: A Domain-Specific LLM Framework for Earth Intelligence}
\author{
 \textbf{Àlex R. Atrio\textsuperscript{1,*,\textdagger}},
 \textbf{Antonio Lopez\textsuperscript{1,*}},
 \textbf{Jino Rohit\textsuperscript{1,*}},
 \\
 \textbf{Yassine El Ouahidi\textsuperscript{2}},
 \textbf{Marcello Politi\textsuperscript{1}},
 \textbf{Vijayasri Iyer\textsuperscript{1}},
 \\
 \textbf{Umar Jamil\textsuperscript{2}},
 \textbf{Sébastien Bratières\textsuperscript{1, 3}},
  \textbf{Nicolas Longépé\textsuperscript{4}}
\\
\\
 \textsuperscript{1}Pi School,
 \textsuperscript{2}Mistral AI,
 \textsuperscript{3}Translated,
 \textsuperscript{4}ESA \(\Phi\)-lab}

\usepackage{booktabs}
\usepackage{tcolorbox}
\usepackage{cuted}
 \usepackage{multirow} 
\usepackage{amsmath}
\usepackage{arydshln}
\usepackage{float}
\usepackage{enumitem}
\usepackage{pgfplots}
\pgfplotsset{compat=1.18}

\definecolor{main}{HTML}{006762}
\definecolor{pretrain}{RGB}{230, 240, 250}
\definecolor{finetune}{RGB}{250, 240, 230}
\definecolor{headerblue}{RGB}{70, 130, 180}
\definecolor{headerorange}{RGB}{210, 105, 30}

\begin{document}
\maketitle

\begin{abstract}

We introduce \textbf{Earth Virtual Expert (EVE)}, the first open-source, end-to-end initiative for developing and deploying domain-specialized LLMs for Earth Intelligence. At its core is \texttt{EVE-Instruct}, a domain-adapted 24B model built on Mistral Small 3.2 and optimized for reasoning and question answering. On newly constructed Earth Observation and Earth Sciences benchmarks, it outperforms comparable models while preserving general capabilities.
We release curated training corpora and the first systematic domain-specific evaluation benchmarks, covering MCQA, open-ended QA, and factuality. EVE further integrates RAG and a hallucination-detection pipeline into a production system deployed via API and GUI, supporting 350 pilot users so far.
All models, datasets, and code are ready to be released under open licenses as contributions to our field at \href{https://huggingface.co/eve-esa}{huggingface.co/eve-esa}  and \href{https://github.com/eve-esa}{github.com/eve-esa}.

% NL: "Earth Intelligence is the fusion of Earth Observation, Earth system science, and geospatial context into AI systems capable of perceiving, reasoning about, and predicting the dynamics of our planet, enabling informed decisions at every scale from satellite operations to global policy."

\end{abstract}

\footnotetext[1]{Equal contribution.}
\footnotetext[2]{Corresponding author: \href{mailto:alex.atrio@picampus-school.com}{alex.atrio@picampus-school.com}}
\renewcommand{\thefootnote}{\arabic{footnote}}

% SUBMISSION STUFF

% We introduce Earth Virtual Expert (EVE), the first open-source, end-to-end initiative for developing and deploying domain-specialized LLMs for Earth Intelligence. At its core is EVE-Instruct, a domain-adapted 24B model built on Mistral Small 3.2 and optimized for reasoning and question answering. On newly constructed Earth Observation and Earth Sciences benchmarks, it outperforms comparable models while preserving general capabilities.
% We release curated training corpora and the first systematic domain-specific evaluation benchmarks, covering MCQA, open-ended QA, and factuality. EVE further integrates RAG and a hallucination-detection pipeline into a production system deployed via API and GUI, supporting 350 pilot users.
% All models, datasets, and code are publicly available at huggingface.co/eve-esa and github.com/eve-esa.

% We introduce Earth Virtual Expert (EVE), an open-source Earth Intelligence system centered on EVE-Instruct, a domain-specialized LLM that outperforms comparable models on domain-specific benchmarks without degrading general capabilities. We release curated training and benchmarking data, and integrate retrieval grounding and hallucination detection into a deployed system.

% Domain-Specific Language Model, LLM, Earth Intelligence, Retrieval-Augmented Generation, Hallucination

% ====================================================

\section{Introduction}
\label{sec:intro}

Earth Observation (EO) (the acquisition of information about Earth's physical, chemical, and biological systems using remote sensing technologies, primarily satellites) and Earth Sciences research generates vast amounts of high‑value knowledge.
% spanning scientific publications, technical reports, datasets, models, and analytical tools.
Yet this knowledge remains fragmented across heterogeneous sources and formats, creating a significant barrier for potential users such as practitioners and decision makers. It also remains crucial for experts in Earth Observation and the wider Earth sciences to continually broaden their understanding across adjacent fields, given the increasingly interdisciplinary nature of modern environmental challenges.
% Those affected by this fragmentation include, inter alia, remote‑sensing scientists, Earth system researchers, environmental analysts, mission operators and decision‑makers, AI/ML practitioners entering the Earth domain, and the growing community of non‑expert or cross‑disciplinary users. 
Accessing
% and integrating 
this information typically requires deep expertise,
% in EO physics and thematic knowledge across Earth system domains,
% further limiting the ability to develop a comprehensive understanding of highly specialized topics.
limiting comprehensive understanding.
% 
% This fragmentation also contributes to a persistent trust and usability gap: decision-makers increasingly rely on EO‑derived information, but they require transparent, explainable, and scientifically robust answers that traditional systems struggle to deliver \cite{KNUTTI201921}.
% This fragmentation creates a trust gap: decision‑makers require transparent, scientifically-robust EO insights that traditional systems struggle to provide \cite{KNUTTI201921}. As the EO community transitions to Earth Action~\footnote{referring to the ability of decision systems to support environmental decisions, operations, or interventions.}, there is an increasing need for systems that do more than access information. They must be capable of interpreting it, contextualizing it within geophysical and socio‑environmental processes, and reasoning across heterogeneous data sources. Earth Intelligence (EI) provides this cognitive layer, enabling a deeper understanding of the Earth system and supporting informed, timely, and reliable action.
% 
This fragmentation creates a trust gap: decision-makers require transparent and scientifically robust EO and Earth Sciences insights that traditional systems struggle to provide \cite{KNUTTI201921}. As the community moves toward Earth Action, the ability of decision systems to support environmental decisions and interventions, there is growing demand for systems that not only retrieve information, but interpret and reason across heterogeneous sources. Earth Intelligence (EI) aims to provide this integrative reasoning layer to support informed and reliable decision-making.

% Recent advances in Large Language Models (LLMs) offer a promising mechanism for natural-language interaction with complex knowledge ecosystems. Yet, general‑purpose models lack the domain specificity and scientifically grounded evaluation protocols required for reliable use in Earth Intelligence applications. Addressing these limitations requires an end-to-end approach that combines domain-adaptive training, retrieval-based grounding, reliability mechanisms, and real-world deployment.

Recent advances in LLMs enable natural-language interaction with complex knowledge ecosystems, yet general-purpose models lack the domain specificity and rigorous evaluation required for reliable EI applications. Addressing this gap requires an end-to-end approach combining domain adaptation, retrieval grounding, reliability mechanisms, and deployment.
% We address these challenges with EVE, an open and modular suite designed for the EO and Earth Sciences communities, and more broadly for advancing the emerging field of EI. EVE is developed within an initiative of the European Space Agency (ESA) \(\Phi\)-lab and has been deployed in a pilot setting, serving 350 users over 6 months. The system integrates multiple document collections, including public encyclopedic resources, ESA documentation, and scientific publisher content. This enables grounded responses across a broad range of EO and Earth science domains, such as remote‑sensing physics, geophysical retrievals, Earth system processes, environmental monitoring, and geospatial analytics. Our contributions are:
 
We introduce EVE, an open and modular framework for EO and Earth Sciences, developed within an initiative of the European Space Agency (ESA) \(\Phi\)-lab and deployed in a six-month pilot serving 350 users. The system integrates heterogeneous knowledge sources, including encyclopedic, institutional, and scientific publisher content, enabling grounded reasoning across diverse EO and Earth Sciences domains. Our contributions are:

\begin{enumerate}[noitemsep, topsep=0pt]
    \item \texttt{EVE-Instruct}: a specialized 24B LLM for EI.
    \item A curated EO and Earth Sciences corpus (2.8B tokens) and large-scale synthetic instruction dataset (10.7B total tokens).%: 20.9M input tokens, 60.1M output tokens, and 10.6B context tokens).
    \item The first manually created EO and Earth Sciences evaluation benchmarks (5693 samples) covering diverse forms of Question-Answering (QA) and factuality.
    \item A deployed RAG- and hallucination-aware chat system accessible via GUI and API.
    \item Open release of models, datasets, and code to support reproducible domain-specific LLM development.
\end{enumerate}

In our experiments, \texttt{EVE-Instruct} consistently outperforms comparable models in its size range on our specific benchmarks, while preserving general capabilities, demonstrating that carefully engineered domain-specific systems can achieve strong practical performance without substantially increasing model size.

% ====================================================
% ====================================================
\begin{figure*}[!ht]
    \centering
    \includegraphics[width=1\linewidth]{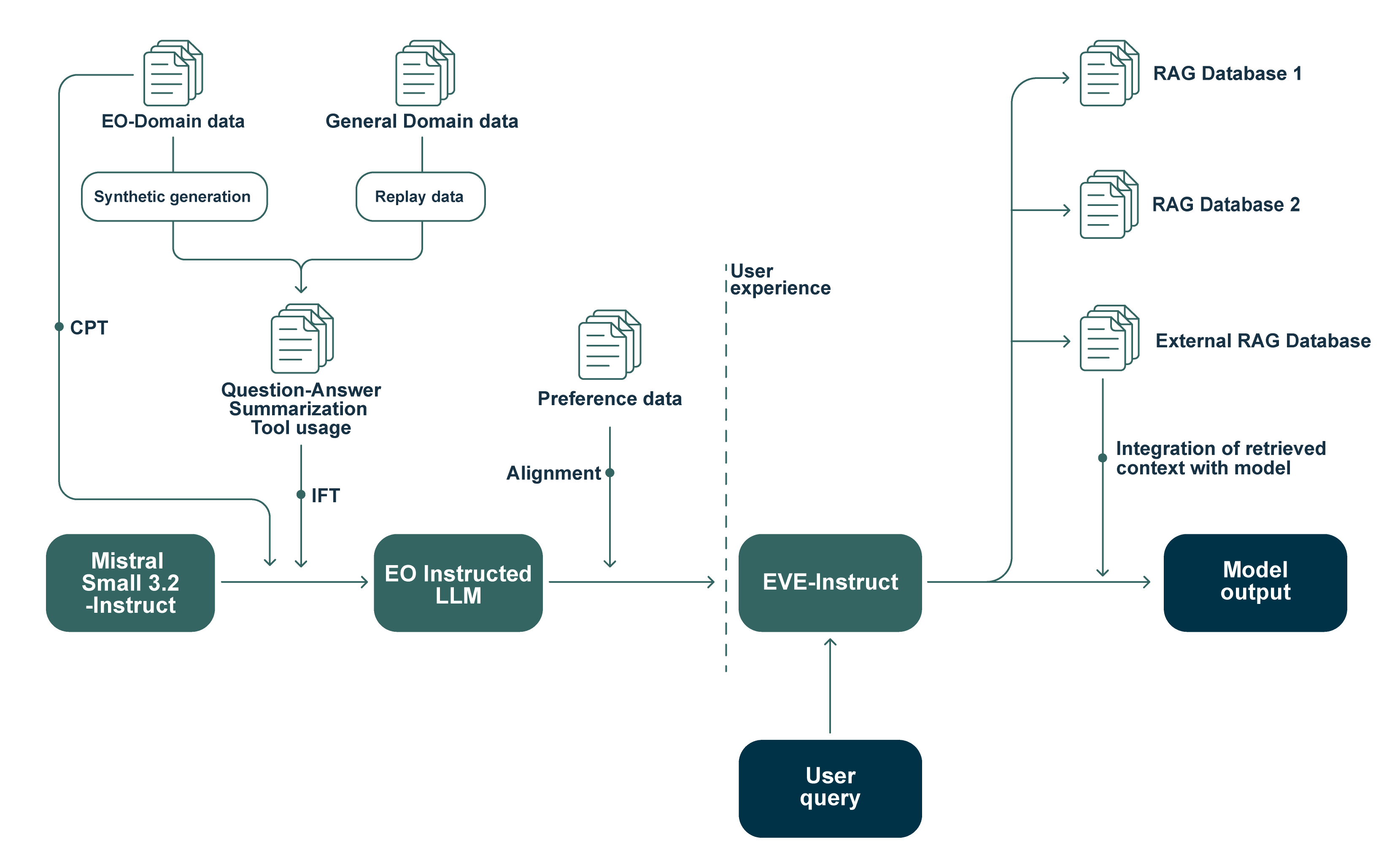}
    \caption{System architecture of EVE depicting component interactions.}
    \label{fig:pipeline}
\end{figure*}
% ====================================================
% ====================================================

\section{Related Work}
\label{sec:related}

% Recently, LLMs tailored for specific fields such as law, medicine, finance, and science have gained significant traction, often matching or surpassing general-purpose models. Developing these specialized models requires careful attention to several factors, including the curation of high-quality domain-specific datasets, design of effective pretraining strategies, mechanisms to detect and mitigate hallucinations, and the creation of practical systems that effectively incorporate domain knowledge.

Recently, domain-specialized LLMs are increasingly achieving performance comparable to or exceeding that of general-purpose models, contingent upon several parts of the system design.
% 
% \textbf{Data Collection and Pretraining.} 
% Large-scale pretraining corpora differ in their design tradeoffs between diversity, scale, filtering rigor, and reproducibility.
% The Pile~\cite{thepile} integrates 22 heterogeneous sources to maximize domain coverage.
% RedPajama-V2 ~\cite{redpajama} emphasizes scale and flexible quality control.
% % curated ~30T tokens of web data annotated with extensive quality signals, enabling customizable filtering rather than enforcing a fixed pipeline. 
% Dolma~\cite{soldaini2024dolma} focuses on reproducibility by providing the complete preprocessing and deduplication toolkit.
% % alongside its 3T-token corpus.
% % Fineweb~\cite{fineweb} applies empirically optimized filtering and large-scale deduplication to produce a 15T-token high-quality web corpus, with FineWeb-Edu (1.3T tokens) further isolating instructional content via LLM-based scoring to enhance reasoning performance.
% FineWeb~\cite{fineweb} focuses on optimized filtering and large-scale deduplication, with an additional subset of instructional content.
% Together, these efforts illustrate complementary strategies for constructing high-quality general-domain corpora, which inform but do not directly address the challenges of domain-specific language modeling.
% 
% \textbf{Data Collection and Pretraining.}
Large-scale pretraining corpora reflect different tradeoffs in diversity, scale, filtering, and reproducibility. The Pile~\cite{thepile} integrates heterogeneous sources to maximize coverage, RedPajama-V2~\cite{redpajama} emphasizes scale and flexible quality control, Dolma~\cite{soldaini2024dolma} prioritizes reproducible preprocessing, and FineWeb~\cite{fineweb} focuses on large-scale filtering and deduplication, including an instructional subset. Together, these efforts advance general-domain data curation but do not directly address the challenges of domain-specific language modeling.

% \textbf{LLMs for Earth Sciences.}
In scientific and EO or Earth Sciences domains, corpus design and domain-adaptive pretraining is central to performance. INDUS~\cite{bhattacharjee2024indus} and K2~\cite{deng2024k2} show that curated scientific corpora and continuous pretraining strengthen domain fidelity and reasoning, while 
% Indus~\cite{bhattacharjee2024indus} demonstrates that curated multidisciplinary scientific corpora improve domain fidelity across the physical sciences, while K2~\cite{deng2024k2} shows that continued pretraining on geoscience literature and structured resources strengthens domain-specific reasoning.
AstroLLaMA~\cite{nguyen2023astrollama}, AstroMLab~\cite{de2024astromlab}, and COSMOSAGE~\cite{de2025cosmosage} demonstrate similar gains by training on scientific publications and observational data, including at compact model scales.

Beyond corpus adaptation, recent work explores spatial reasoning and geospatial system integration. GeoLLM~\cite{manvi2023geollm}
% shows that LLMs encode implicit geographic knowledge that can be enhanced through grounding with structured geodata such as OpenStreetMap.
shows that LLMs encode geographic knowledge that can be enhanced through grounding with structured geodata.
Frameworks like GeoGPT~\cite{zhang2024geogpt} and BB-GeoGPT~\cite{zhang2024bb} combine LLMs with GIS toolchains for spatial analysis, while agent-based approaches such as GeoAgent~\cite{geoagent}, UrbanGPT~\cite{urbangpt}, and ChatGeoAI~\cite{mansourian2024chatgeoai} enable
% extend this paradigm to
autonomous
% spatial reasoning
and conversational geospatial reasoning.

Several frameworks address hallucination evaluation in LLMs. FEVER~\cite{thorne2018fever}, TruthfulQA~\cite{lin2021truthfulqa}, and HaluEval~\cite{li2023halueval} provide fact-verification and truthfulness benchmarks, including different hallucination types. LLM-Oasis~\cite{scire2025truth} introduces a large-scale benchmark for end-to-end factuality evaluation. SelfCheckGPT~\cite{manakul2023selfcheckgpt} proposes reference-free hallucination detection, while RARR~\cite{gao2023rarr} reduces factual errors through retrieval-based attribution and revision.

%The Pleias-RAG \cite{langlais2025even} model family address this by introducing a new series of models specialized into natively supporting citations and grounding by literal quotes.

% Domain-specific hallucination detection efforts include FactScore~\cite{min2023factscore} for biographical information, MedHalt~\cite{medhalt} for medical claims, and ClimateFever~\cite{diggelmann2020climatefever} for climate science statements. In scientific domains, SciFact~\cite{wadden2020scifact} evaluates citation verification, while numerical reasoning benchmarks like NumGLUE~\cite{mishra2022numglue} assess quantitative accuracy—particularly important in geospatial analysis where incorrect coordinates, measurements, or calculations can lead to flawed analyses or decision-making.

% \paragraph{Retrieval-Augmented Generation.} RAG approaches~\cite{lewis2020retrieval, gao2023rag, izacard2021fewshot} offer a complementary strategy to reduce hallucinations and improve factual grounding by allowing models to retrieve and condition on relevant external knowledge during generation. Domain-specific RAG systems include Atlas~\cite{izacard2022atlas} for knowledge-intensive tasks, GopherCite~\cite{menick2022gophercite} for attributed generation, and Toolformer~\cite{schick2023toolformer} for teaching models to use external tools. In geospatial contexts, combining retrieval with structured geodata repositories and spatial databases can provide verifiable grounding for location-based queries and spatial reasoning tasks.

% ====================================================
\section{EVE System Overview}
\label{sec:system}

% The full, deployed EVE system is composed of several self-contained components that work together to provide a factual answer with the corresponding knowledge used to generate it. Their interaction is illustrated in Figure~\ref{fig:pipeline}:
The deployed EVE system consists of modular components that jointly generate grounded responses (Figure~\ref{fig:pipeline}):
\begin{itemize}[noitemsep,topsep=0pt]
    % \item \textbf{\texttt{EVE-Instruct}} serves as the core language model and is invoked by the other components for different tasks including: query rewriting, answer generation, or conversation summarization.
    % \item \textbf{Knowledge bases} represent the main source of information to ground the responses of the model, as described in Section~\ref{sec:grounded}. Five main sources are available, all featuring trusted or peer-reviewed sources of EI knowledge:
    % An open-access curated database with $\sim$270k documents,
    % a closed-access database with $\sim$70k documents,\footnote{Access provided under a formal partnership agreement with Wiley. See \href{https://www.wiley.com}{wiley.com}.}
    % an open-access database comprised of various ESA documents ($\sim$10k),
    % and the private users' collections.    
    % All support hybrid retrieval, enabling both semantic search and metadata filtering.
    % \item \textbf{Retrieval Pipeline}: retrieves the most relevant documents for a given user query based on content and search parameters (see Section~\ref{par:retrieval-pipeline}).
    % \item \textbf{Chat System}: This component manages the conversation flow, memory, and context allocation. Its goal is to maintain awareness of prior interactions while preserving the capacity of the context window for the current turn (see Section \ref{sec:chat_system}). 
    \item \textbf{\texttt{EVE-Instruct}}: core LLM for answer generation, query rewriting, and summarization (Section~\ref{sec:training}).
    \item \textbf{Knowledge Bases (KB)}: curated domain-specific sources (open-access, proprietary,\footnote{Provided under a partnership agreement with \href{https://www.wiley.com}{Wiley}.} 
    ESA documents, and private collections) totaling $\sim$365k documents, supporting hybrid semantic and metadata retrieval (Section~\ref{sec:grounded}).    
    \item \textbf{Retrieval Pipeline}: selects relevant documents based on query content and filters (Section~\ref{par:retrieval-pipeline}).
    \item \textbf{Chat System}: manages dialogue state, memory, and context allocation (Appendix~\ref{sec:appendix-architecture}).
\end{itemize}

% ====================================================

\section{EO and Earth Sciences Text Corpus}
\label{sec:data}

% \textbf{Data Collection.}
We curate a large-scale EO and Earth Sciences corpus by manually selecting 172 sources spanning 22 trusted publishing institutions (see Table~\ref{tab:data_distribution} in Appendix~\ref{sec:appendix-corpus-creation}) using a custom scraping framework. The final corpus is comprised of 5.3B tokens: 4.2B from open-access sources and 1.1B from Wiley proprietary content (see Section~\ref{sec:system}). We use filtered subsets of this corpus for synthetic data generation (Section~\ref{sec:synthetic-data}) and RAG (Section~\ref{sec:grounded}). We publicly release 2.8B tokens of the open-access portion in due consideration of licensing conditions, as detailed in Appendix~\ref{sec:appendix-corpus-creation}.

% \textbf{Data Processing Pipeline.}
Our data processing pipeline transforms raw documents into clean, structured training data. First, we extract machine-readable text from the original files with Trafilatura~\cite{barbaresi-2021-trafilatura} for HTML, and Nougat~\cite{blecher2023nougat} OCR for PDFs, selected after benchmarking for \LaTeX-based formula and table extraction.
% Plain text formats do not require any specialized processing.
% 
% \paragraph{Deduplication.}
We apply SHA-256 based hashing at file level to remove exact duplicates and MinHash LSH\footnote{\href{https://github.com/ekzhu/datasketch}{github.com/ekzhu/datasketch}} to remove near-duplicate text segments within the file itself.
%
% \paragraph{Cleaning.} 
%
% This stage applies OCR error correction to remove any artifacts or erraneous extraction during the oCR phase, a LaTeX correction module using an LLM and a few handwritten regex rules.
%
% (or this version?)
% We apply a lightweight post-processing step to correct common OCR and parsing errors in the extracted scientific documents. This hybrid approach first applies a simple rule based normalization step to remove the most frequent noise patterns we observed, then feeds the cleaned text to a lightweight LLM based LaTeX correction module which repairs malformed LaTeX commands, mismatched environments, and other syntax errors while carefully preserving the document’s mathematical structure.
We perform lightweight post-processing to correct OCR noise and malformed \LaTeX \ using rule-based normalization and an LLM-based syntax repair module.
%
%\paragraph{PII Removal.}
We use Microsoft Presidio\footnote{\href{https://github.com/microsoft/presidio}{github.com/microsoft/presidio}} with Flair \texttt{ner-english-large} model~\cite{akbik2019flair} to anonymize author names as \texttt{[AUTHOR]} and emails as \texttt{[EMAIL]} to ensure removal of personally identifiable information.
%
%\paragraph{Metadata Extraction.}
We extract structured metadata (e.g., DOI, URL, title, journal) for academic PDFs by identifying DOIs via regular expressions and querying CrossRef API.\footnote{\href{https://www.crossref.org/}{crossref.org}}
% The pipeline yields metadata for 290k records, with missing rates of 4.51\% for authors, 3.96\% for journal, and below 0.1\% for all other fields.

\section{EO and Earth Sciences Benchmark}

\label{subsec:data-eval}
Due to the lack of standardized benchmarks for dialogue and NLP capabilities in EO and Earth Sciences, we construct manually curated evaluation sets targeting domain-relevant tasks (Table~\ref{tab:eve-tasks}). To our knowledge, these constitute the first systematic benchmarks within the domain for language modeling. The datasets are built in two stages: candidate samples are generated by both humans and LLMs, and subsequently reviewed and refined by independent human annotators. We recruited 25 EO and Earth Sciences experts as human annotators and provided them with annotation guidelines.
Table~\ref{tab:benchmark-examples} provides representative examples from each category.

\begin{table}[htbp]
\small
\centering
\begin{tabular}{llllr l}
\toprule
Task Type & Samples \\
\midrule
MCQA (Multiple Answer) & 431  \\
MCQA (Single Answer) & 1261  \\
Hallucination Detection & 2326 \\
Open-Ended QA (No context) &  1257  \\
Open-Ended QA (with Context) & 418  \\
\bottomrule
\end{tabular}
\caption{EO and Earth Sciences benchmark suite.
Multiple Choice Question Answering (MCQA) differ in the number of correct options.
Hallucination detection is a balanced binary QA classification task.
Open-ended QA evaluate self-contained and context-grounded questions.
}
\label{tab:eve-tasks}
\end{table}

\begin{table*}[!htb]
\small
\centering
\begin{tabular}{@{}p{2.1cm}p{12.8cm}@{}}
\toprule
\textbf{Task Type} & \textbf{Example} \\
\midrule
MCQA

(Multiple

Answer) &
\textbf{Q:} Which statements are true about surging glaciers? \newline
\hspace*{1em}A) Surging glaciers are subject to cyclical flow instabilities \newline
\hspace*{1em}B) Surging glaciers flow at a constant rate down a mountain \newline
\hspace*{1em}C) [\ldots] a glacier's velocity moves up to 100 times faster than normal \newline
\hspace*{1em}D) Surging glaciers suddenly increase in mass \newline
$\rightarrow$ \textbf{Answers: A, C} \\
\midrule
MCQA

(Single

Answer) &
\textbf{Q:} What is the main obstacle to using optical EO in the aftermath of a severe weather event? \newline
\hspace*{1em}A) Sensor calibration issues \newline
\hspace*{1em}B) Atmospheric interference \newline
\hspace*{1em}C) Low light conditions \newline
\hspace*{1em}D) Cloud cover \newline
$\rightarrow$ \textbf{Answer: D} \\
\midrule
Hallucination Detection &
\textbf{Q:} What advantages do radar systems offer for environmental monitoring compared to optical sensors? \newline
\textbf{Hallucinated Answer:} Radars are active sensors that create their own electromagnetic signals.\ \underline{Radars cannot operate at night and are ineffective in hazy or smoky environments.}\ SAR [\ldots] uses the movement between the antenna and the target area [\ldots] to create high-resolution images for remote sensing. \newline
\textbf{Correct Answer:} Radars are active sensors [\ldots] Radars can operate day and night, and penetrate clouds, haze, and smoke. SAR [\ldots] \newline
$\rightarrow$ \textbf{Label: Hallucinated} (\underline{underlined} span) \\
\midrule
Open-Ended QA &
\textbf{Q:} What is the calving front of an ice stream? \newline
\textbf{A:} The point where icebergs are lost from a glacier. \\
\midrule
Open-Ended QA w/ Context &
\textbf{Q:} How does LiDAR map buildings? \quad \textit{[3 source documents provided as context]} \newline
\textbf{A:} LiDAR maps buildings by sending laser pulses that reflect off rooftops, walls, and other structural surfaces, recording precise 3D coordinates. These points are stored in point clouds [\ldots] \\
\bottomrule
\end{tabular}
\caption{Representative examples from each evaluation category in the EO and Earth Sciences benchmark suite (see Table~\ref{tab:eve-tasks}). In hallucination detection, \underline{underlined} spans are the annotated hallucinated content.}
\label{tab:benchmark-examples}
\end{table*}

% ====================================================

\section{Model Fine-tuning}
\label{sec:training}

Adapting an instruction-tuned LLM to our target domain requires incorporating domain-specific knowledge without degrading its instruction-following, conversational stability, or tool-use behavior. The 5.3B tokens of EO and Earth Sciences text (Section~\ref{sec:data}) are sufficient to consider full-parameter fine-tuning over LoRA~\cite{hu2022lora}, but insufficient for standalone continuous pretraining (CPT)~\cite{gururangan2020don,zixuan2023continual}. In preliminary experiments, the higher learning rates required for factual acquisition led to degradation of instruction-following behavior. To address this trade-off, we adopt a fine-tuning strategy that interleaves instruction fine-tuning (IFT) and long-form text, each mixing general-domain replay data with synthetic EO and Earth Sciences text. This enables domain adaptation while preserving general interactive capabilities. Both training and synthetic data generation are performed using an internal training framework. %Mistral AI Forge, a Mistral internal training framework.

% To address this trade-off, we adopt a hybrid training strategy combining (i) supervised fine-tuning (SFT) on synthetic instruction data with (ii) pretraining-style raw-text supervision. Domain text is reformulated into instruction-following examples and complemented with synthetic raw text and replay data from the original training distribution. This enables domain adaptation while preserving general interactive capabilities.
% 

% Both training and synthetic data generation are performed using Mistral AI Forge, Mistral’s internal model development platform.

\subsection{Fine-tuning Data Synthesis}
\label{sec:synthetic-data}

% Our synthetic supervision 
Our fine-tuning data consists of two components:  long-form text and instruction-formatted text. 
% We use our corpus (Section~\ref{sec:data}) for both CPT and for generating instruction-formatted synthetic data.
Detailed distributions are provided in Table~\ref{tab:data_mix}.
Due to licensing conditions of source materials, we publicly release a curated subset of 10.7B tokens (20.9M input, 60.1M output, and 10.6B context tokens), of the full dataset used for training.

\textbf{Long-form text.} 
%We include EO and Earth Sciences synthetic raw text in the raw-text batches, allowing domain knowledge to be integrated alongside IFT. 
Our long-form fine-tuning data consists of long-form general-domain replay data alongside EO and Earth Sciences long-form text, which in turn consists of (i) a small portion of EO and Earth Sciences raw text from our corpus, and (ii) synthetically generated EO and Earth Sciences text. The former consists of either random samples from the corpus (Raw) or high-quality filtered samples (Best Chunks). The latter is generated with an Active Reading~\cite{lin2025learning} pipeline, which reorganizes salient content to concentrate factual information and reinforce terminology, using either task-specific or predefined strategies. Strategy selection is performed by Mistral Medium 3.1, while Mistral Small 3.2 performs generation to maintain distributional alignment with the base model, as advised in~\citet{lin2025learning}.

\textbf{Instruction-formatted text.}
EO and Earth Sciences documents from our corpus (Section~\ref{sec:data}) are transformed into instruction–response pairs, including:
(i) contextual and non-contextual Question Answering (QA) (ContextQA, SelfQA), 
(ii) long and multi-document QA (LongQA), 
(iii) multi-hop QA~\cite{shen2025hopweaver},
(iv) self-referential alignment prompts (role, developer, and capability specification).
We use various high-quality models for generation, including: Mistral Large 3, GPT-4o Mini, Mistral Medium 3.1, Qwen3-235B, DeepSeek-R1, DeepSeek v3.1, Qwen2.5-72B. This is mixed with instruction-formatted replay data.
%(iv) refusal examples,
%(v) summarization tasks, and
% All samples are grounded in source documents. Generation and filtering use distinct models , with automated quality control removing hallucinated or weakly grounded samples. 

\textbf{Quality control and filtering.}
We use LLM-based judges to assess domain relevance, factual quality and grounding. Long-form and Instruct text filtering uses either of the larger models used for generation as listed just above, with judge labels (``Best'', ``Good'' or ``Bad'') determining retained samples. In total, we generate approximately 21B tokens of synthetic data, from which a filtered subset forms the synthetic pool used in the final training mixture (Table~\ref{tab:data_mix}).

\begin{table}[h]
    \centering
    \scriptsize
    \resizebox{\columnwidth}{!}{
    \begin{tabular}{@{}lr|lr@{}}
        \toprule
        \multicolumn{2}{c|}{Long-form text: 30\% (10B)} &
        \multicolumn{2}{c}{Instruction-formatted data: 70\% (23.5B)} \\
        \midrule
        %Mistral 
        Long-Form Replay & 50\% & 
        %Mistral 
        Instruct-formatted Replay  & 60\% \\
        % \rowcolor{gray!10}
        \textbf{EO and Earth Sciences} & \textbf{50\%} & \textbf{EO and Earth Sciences} & \textbf{40\%} \\
        \quad Raw & 2\% & \quad ContextQA (Best) & 12\% \\
        \quad Best Chunks & 14\% & \quad ContextQA (Good) & 21\% \\
        \quad Active Reading (Agnostic) & 28\% & \quad SelfQA & 2.6\% \\
        \quad Active Reading (Specific) & 6\% & \quad MultiHop QA & 2.1\% \\ 
         & & \quad Long QA & 2.6\% \\
        \bottomrule
    \end{tabular}
    }
    \caption{Distribution of training data across long-form and instruction-formatted data.}
    \label{tab:data_mix}
\end{table}

\subsection{Fine-tuning mixing long-form and instruction text} 

We fine-tune Mistral Small 3.2 (24B parameters, 128k context) interleaving instruction-formatted with long-form text within the same training runs. 
Increasing the proportion of EO and Earth Sciences data improves domain-specific benchmarks, but comes at the cost of reduced performance on general capabilities, particularly tool usage and structured reasoning. Replay data mitigates catastrophic forgetting and stabilizes interaction behavior.
To address this trade-off, during fine-tuning, we vary
(i) the ratio of long-form versus instruction-formatted text and (ii) the proportion of general domain replay versus domain-specific data within each type, so that the percentages presented in Table~\ref{tab:data_mix} are cumulative ratios taken over the entire fine-tuning process.
Further, we use a learning rate schedule intermediate between typical IFT and CPT settings to balance factual integration and alignment stability. Finally, since runs trained with different data mixtures trade off domain and general performance differently, we merge checkpoints from ten runs using uniform parameter interpolation.

Our choice of replay-based training with checkpoint merging is motivated by stability and scalability in a production setting. We explored alternatives including LoRA~\cite{hu2022lora} and regularization-based methods during development, but found that interleaved replay with checkpoint merging offered the best trade-off between domain acquisition and capability retention at our training scale. Recent work has proposed complementary strategies such as self-synthesized rehearsal~\cite{huang2024mitigating} and selective parameter freezing~\cite{hui2025hft}, which may offer further improvements and constitute promising directions for future work.

%\subsection{Checkpoint Merging}

%Different training configurations exhibit complementary strengths across evaluation regimes. Rather than selecting a single checkpoint, we merge selected checkpoints from ten training runs exploring different data mixtures and supervision balances.

%Checkpoint merging is performed with uniform parameter interpolation. % with tuned weights.%, allowing each selected checkpoint to contribute to the final model. This allows us to combine complementary behaviors observed across runs. Empirically, merged models provide improved trade-offs between domain specialization and general capabilities relative to individual checkpoints.

\subsection{Alignment}

We apply Online Direct Preference Optimization~\cite{qi2024online} as a final alignment stage to refine formatting, stylistic consistency, and preference adherence. We follow the same alignment recipe and preference training data as in~\citet{liu2026ministral}. This final stage improves formatting consistency and adherence to preference signals, while preserving domain knowledge acquired during earlier training.

% ====================================================

\section{Evaluation}
\label{sec:evaluation}

\textbf{Setup.} We evaluate \texttt{EVE-Instruct} on both domain-specific benchmarks (Section~\ref{subsec:data-eval}) and general-domain benchmarks to assess domain gains and preservation of general capabilities. We compare against the parent model and four additional LLMs of comparable scale ($\sim$24B parameters).

% \begin{table*}[!htb]
% \centering
% \small
% \setlength{\tabcolsep}{3pt}   % default is 6pt
% % \begin{tabular}{llcccccccc}
% \begin{tabular}{@{}llcccccccc@{}}
% \toprule
%  & & \multicolumn{2}{c}{MCQA Multiple} & MCQA Single & Hallucination & \multicolumn{2}{c}{Open-Ended} & \multicolumn{2}{c}{Open-Ended w/ Context} \\
% \cmidrule(lr){3-4}
% \cmidrule(lr){5-5}
% \cmidrule(lr){6-6}
% \cmidrule(lr){7-8}
% \cmidrule(lr){9-10}
% Model & Size (B) & IoU & Acc. & Acc. & F1 & Judge & EVE WR & Judge & EVE WR \\
% \midrule
% Llama4 Scout & 109-A17 & 80.32 & 71.23 & 91.67 & 66.08 & 87.37 & 53.95 & 71.73 & 58.31 \\
% Qwen3 & 30-A3 & 78.40 & 66.36 & 93.02 & 81.30 & 94.92 & 50.70 & \textbf{81.81} & 52.12 \\
% Gemma3 & 27 & 73.60 & 57.54 & 87.31 & 75.07 & 94.41 & 50.92 & 78.31 & 51.58 \\
% Mistral Small 3.2 & 24 & 80.19 & 70.30 & 83.51 & 82.19 & 91.78 & 51.69 & 71.93 & 57.27 \\
% \hdashline
% \texttt{EVE-Instruct} & 24 & \textbf{86.12} & \textbf{77.73} & \textbf{96.35} & \textbf{84.70} & \textbf{96.40} & --- & 78.28 & --- \\
% \bottomrule
% \end{tabular}
% \caption{Model performance across EO and Earth Sciences benchmark tasks presented in Table~\ref{tab:eve-tasks} (0-shot). EVE WR (win rate) indicates percentage of pairwise comparisons where \texttt{EVE-Instruct} is preferred over the comparison model ($>50\%$ means is EVE preferred).}
% \label{tab:eo-benchmark-results-0-shot}
% \end{table*}

\begin{table*}[!htb]
\centering
\small
\setlength{\tabcolsep}{3pt}
\begin{tabular}{@{}llccccccccc@{}}
\toprule
 & & \multicolumn{2}{c}{\small MCQA Mult.} & {\small MCQA Sing.} & {\small Hallucin.} & \multicolumn{2}{c}{\small Open-Ended} & \multicolumn{2}{c}{\small Open-Ended w/ Context} & \\
\cmidrule(lr){3-4}
\cmidrule(lr){5-5}
\cmidrule(lr){6-6}
\cmidrule(lr){7-8}
\cmidrule(lr){9-10}
{\small Model} & {\small Size (B)} & {\small IoU} & {\small Acc.} & {\small Acc.} & {\small F1} & {\small Judge} & {\small EVE WR} & {\small Judge} & {\small EVE WR} & {\small Rank $\downarrow$} \\
\midrule
Llama4 Scout & 109-A17 & 80.32 & 71.23 & 91.67 & 66.08 & 87.37 & 53.95 & 71.73 & 58.31 & 3.67 \\
Qwen3 & 30-A3 & 78.40 & 66.36 & 93.02 & 81.30 & 94.92 & 50.70 & \textbf{81.81} & 52.12 & 2.67 \\
Gemma3 & 27 & 73.60 & 57.54 & 87.31 & 75.07 & 94.41 & 50.92 & 78.31 & 51.58 & 3.83 \\
Mistral Small 3.2 & 24 & 80.19 & 70.30 & 83.51 & 82.19 & 91.78 & 51.69 & 71.93 & 57.27 & 3.50 \\
\hdashline
\texttt{EVE-Instruct} & 24 & \textbf{86.12} & \textbf{77.73} & \textbf{96.35} & \textbf{84.70} & \textbf{96.40} & --- & 78.28 & --- & \textbf{1.33} \\
\bottomrule
\end{tabular}
\caption{Model performance across EO and Earth Sciences benchmark tasks presented in Table~\ref{tab:eve-tasks} (0-shot). EVE WR (win rate) indicates percentage of pairwise comparisons where \texttt{EVE-Instruct} is preferred over the comparison model ($>50\%$ means EVE is preferred). Rank $\downarrow$ (lower is better) reports the average per-metric rank across MCQA multiple (IoU and Accuracy), MCQA single (Accuracy), Hallucination (F1), Open-ended QA (Judge), and Open-Ended QA with Context (Judge).}
\label{tab:eo-benchmark-results-0-shot}
\end{table*}

For open-ended benchmarks, we adopt the LLM-as-a-judge framework~\cite{wang2023evaluating} to evaluate answer correctness. Each candidate response is scored by an LLM judge conditioned on the question, reference answer, and, when applicable, retrieved context, using a 0–5 scale with predefined criteria (Appendix~\ref{sec:appendix-prompts}). To improve robustness and mitigate individual model bias, we aggregate scores from a panel of judges\footnote{Mistral Large 3, GPT-4.1 mini, DeepSeek-V3.2, and Qwen3-235B-A22B}~\cite{verga2024replacing} and report the mean normalized score.
Following \citet{alpaca_eval}, we additionally conduct pairwise preference evaluation (Win Rate), where judges compare two candidate responses and select a winner or tie. The win rate of model $A$ over model $B$ is computed as the average preference across $N$ evaluators:
$\mathrm{WR}_A = \frac{1}{N}\sum_{i=1}^{N}\frac{\mathrm{wins}_{A_i} + 0.5 \cdot \mathrm{ties}_i}{\mathrm{wins}_{A_i} + \mathrm{ties}_i + \mathrm{losses}_{A_i}}$

\textbf{Discussion}. As shown in Table~\ref{tab:eo-benchmark-results-0-shot}, \texttt{EVE-Instruct} achieves the highest performance across both MCQA benchmarks (single- and multiple-answer), indicating effective incorporation of domain-specific knowledge through fine-tuning. On the hallucination detection task, it attains the highest F1 score, reflecting improved discrimination between factual and non-factual responses. \texttt{EVE-Instruct} also leads competing models on open-ended QA without context under both the LLM-as-a-judge and Win Rate evaluations. When retrieval context is provided, Qwen3 achieves the highest LLM-as-a-judge score; however, \texttt{EVE-Instruct} remains competitive and obtains the strongest pairwise preference results, suggesting comparable overall response quality, despite smaller size.

To assess whether domain specialization impacts general capabilities, we report category-level averages across a suite of general-domain benchmarks in Table~\ref{tab:gd-benchmarks}. Each category represents the mean score over multiple underlying benchmarks, whose full breakdown is provided in Appendix~\ref{sec:appendix-evaluation}.
Tool Calling, Instruction Following, and Chat Quality correspond to internal evaluation suites from Mistral.
Across all categories, \texttt{EVE-Instruct} maintains or improves performance with respect to its parent model, indicating that domain-specific adaptation does not degrade general-domain or chat capabilities.
To address the potential overlap between model families used for synthetic data generation (Section~\ref{sec:synthetic-data}) and the LLM-as-a-judge panel, we extend the evaluation with two independent judge families not involved in data generation: Claude Sonnet 4.6 and Gemini 2.5 Flash. As shown in Appendix Table~\ref{tab:benchmark-two-judge}, the resulting rankings are nearly identical (maximum shift: $\pm 0.25$), confirming that the original panel does not exhibit systematic bias favoring \texttt{EVE-Instruct}.

\begin{table}[!htb]
    \centering
    \small
    \setlength{\tabcolsep}{4pt}
    \begin{tabular}{@{}lrrr@{}}
        \toprule
        Category & Small 3.2 & \texttt{EVE-Instruct} & $\Delta$ \\
        \midrule
        Math \& Reasoning        & 50.8 & \textbf{54.9} & +4.1 \\
        Coding                   & 55.6 & \textbf{56.5} & +0.9 \\
        Knowledge                & 67.7 & \textbf{69.0} & +1.3 \\
        Tool Calling             & 87.9 & \textbf{90.9} & +3.0 \\
        Instruction Following    & 80.1 & \textbf{81.2} & +1.1 \\
        Chat Quality             & 90.8 & \textbf{91.7} & +0.9 \\
        \midrule
        Overall         & 72.2 & \textbf{74.0} & +1.8 \\
        \bottomrule
    \end{tabular}
    \caption{General-domain performance after domain adaptation (category-level averages over several standard benchmarks, 0--100 scale).}
    \label{tab:gd-benchmarks}
\end{table}

% ====================================================

\section{Grounded Generation} % RAG + hallucination
\label{sec:grounded}
We developed a RAG pipeline that grounds responses in relevant documents from our KBs (Section~\ref{sec:system}), reducing hallucinations and extending knowledge beyond the training data.

% \paragraph{Vector Database.}
Documents are chunked into $\sim$512-word segments via a two-pass strategy (first by document sections, then by paragraphs or sentences) that preserves \LaTeX \ and Markdown formulas and tables. Uninformative chunks are filtered using statistical heuristics, then enriched with metadata and embedded using \texttt{Qwen3-Embedding-4B} \cite{zhang2025qwen3embeddingadvancingtext}. Embeddings are stored through binary quantization in Qdrant.\footnote{\href{https://qdrant.tech/documentation/guides/quantization/}{qdrant.tech/documentation/guides/quantization/}}
%Documents from the EVE corpus are processed through a multi-stage pipeline. First, documents are chunked into segments of approximately 512 words. To preserve semantic coherence, we apply a two-step chunking strategy: the first pass splits by sections, keeping each section intact; the second pass enforces length constraints by splitting over paragraphs and sentences while ensuring LaTeX and Markdown formulas remain intact.
%Next, uninformative chunks are filtered using heuristics based on statistical text properties like removing chunks with high proportions of anonymized tokens, extreme lengths, and similar artifacts. Finally, each chunk is enriched with metadata from its source document and embedded using \texttt{Qwen3-Embedding-4B} \cite{zhang2025qwen3embeddingadvancingtext}. Embeddings are stored in Qdrant with quantization enabled.\footnote{\href{https://qdrant.tech/documentation/guides/quantization/}{qdrant.tech/documentation/guides/quantization/}}

%\paragraph{Retrieval Pipeline.}
\label{par:retrieval-pipeline}
% The retrieval process begins with the EVE-Instruct query rewrite to adapt user queries to the retrieval scenario. This step transforms the user query into a retrieval-optimized form while integrating conversational context from prior turns.
% For each available knowledge base, the top $K \times 2$ documents are retrieved using embedding similarity. Documents can optionally be filtered by metadata constraints. All retrieved documents are then re-ranked using \texttt{Qwen3-Reranker-4B} \cite{zhang2025qwen3embeddingadvancingtext}, and the top $K$ are returned.
For chunk retrieval, we first apply a query rewriting step using \texttt{EVE-Instruct} by incorporating conversational context, disambiguating, and optimizing retrieval. For each KB, the top $2K$ chunks are retrieved via embedding similarity with optional metadata filtering. The candidates are then re-ranked using \texttt{Qwen3-Reranker-4B} \cite{zhang2025qwen3embeddingadvancingtext}, and the top $K$ documents are selected.
%
%This step expands ambiguous terms that may belong to different domains (e.g., ``ESA'' could refer to the European Space Agency or the Ecological Society of America) and incorporates context from previous conversation turns to resolve references in the current request.

\label{sec:hallucination}
\textbf{Hallucination Detection.}
% The objective of this pipeline is to minimize hallucinations in the system's answer.
To address the issue of factual hallucinations while keeping average answer latency low, we implement a pipeline which starts with hallucination detection and, based on the outcome, optionally proceeds to answer revision.
In the first stage, \texttt{EVE-Instruct} performs fact-checking, acting as an evaluator, and produces a binary hallucination label as well as a justification. If flagged for hallucination, the query is reformulated using the justification to address identified issues using newly retrieved documents. With the retrieved documents, the model generates a revised response, encouraging more conservative and fact-grounded answers. Then, inspired by \citet{ji-etal-2023-towards}, the model critiques the original response using both prior and newly retrieved evidence to produce a revised answer. 
Finally, the model ranks the original and revised outputs based on factuality and supporting evidence, selecting the most reliable response. 

% ====================================================

\section{Deployment}
\label{sec:deployment}

EVE was deployed as a production system supporting 350 users during a six-month pilot from September 2025. The architecture consists of: (i) a single-node Qdrant vector database storing 4.2M dense embeddings with binary quantization; (ii) \texttt{EVE-Instruct}, hosted on RunPod serverless infrastructure with dynamic scaling (1--30 workers) across NVIDIA A100/H100 GPUs; (iii) an Amazon DocumentDB cluster for user management, chat history, and application metadata; (iv) an AWS EC2 backend;\footnote{Instance type t3.large (2 vCPUs, 8\,GB RAM, 320\,GB).} and (v) an AWS CloudFront CDN-managed frontend. A detailed description of the end-to-end system is provided in Appendix~\ref{sec:appendix-architecture}. 

% ====================================================

\section{Conclusion and Future Work}
\label{sec:conclusion}

In this paper, we introduced \textbf{Earth Virtual Expert (EVE)}, an open and modular end-to-end system for building, evaluating, and deploying domain-specialized LLMs for EO and Earth Sciences.
EVE combines (i) large-scale curation and processing, (ii) domain-adapted \texttt{EVE-Instruct} built on Mistral Small 3.2 24B, (iii) domain-specific evaluation benchmarks, and (iv) retrieval-augmented and hallucination-aware grounded generation in a production deployment. 
Across our domain benchmarks, \texttt{EVE-Instruct} improves over other strong models in its parameter range on multiple-choice QA, hallucination detection, and open-ended instruction-following domain specific benchmarks, while remaining competitive on general capabilities.
Beyond offline evaluation, the system has been deployed in a 6-month pilot serving 350 users via GUI and API, demonstrating that domain-specific, grounded scientific assistants can be delivered with practical latency and cost constraints.
We release models, code, curated corpus, manually-created domain benchmarks, and a substantial portion of the synthetically-generated fine-tuning dataset used to create \texttt{EVE-Instruct}.

% As the EO community moves toward Earth Intelligence and Earth Action, there is a growing need to fuse diverse textual knowledge with the large volumes of spaceborne observations, complemented by in‑situ measurements and Earth system models. Recent advances in Geospatial Foundation Models (GFMs), Vision–Language Models (VLMs), and spatially aware embedding spaces provide the computational basis for this joint text–data fusion \cite{11303008}. Together, these components enable shared representations that support reasoning, attribution, scenario exploration, and decision‑making, forming the foundation of a scalable cognitive Earth system capable of integrating observations, models, and expert knowledge into actionable intelligence. 

As EO and Earth Sciences advance toward Earth Action~\cite{esaaction}, there is increasing need to integrate textual knowledge with spaceborne observations, in-situ data, and Earth system models. Recent advances in Geospatial Foundation Models, Vision–Language Models, and spatial embeddings enable joint text–data representations that support reasoning and decision-making \cite{11303008}. Building on this, we aim to extend EVE beyond text into a multimodal, agentic platform capable of reasoning over imagery and geospatial data, and supporting multi-step scientific workflows for large-scale EO and Earth Sciences analysis and data-driven inference.

% Building on this foundation, we aim to evolve EVE from a text‑based system into a fully multimodal, agentic platform for Earth Intelligence. This includes incorporating GFMs and geospatial representations to enable reasoning over imagery, maps, and scientific documents, and extending EVE toward agentic scientific workflows capable of multi‑step planning, execution, large‑scale EO data mining, and deeper, data‑centric scientific inference.

% such as data discovery, cross-repository retrieval, geospatial tool invocation, and structured report generation, while maintaining explicit intermediate reasoning states, provenance tracking, and evaluation of end-to-end task success.

% ====================================================

\section*{Limitations}
\label{sec:limitations}
We highlight key limitations:
(i) Licensing prevents redistribution of 1.1B Wiley tokens ($\sim$21\% of the corpus); we release the open-access subset, pipelines, and synthetic data, so exact reproduction requires independent access to licensed or non-re-distributable content.
(ii) Evaluation coverage remains limited in task diversity and scale, including human evaluation.
(iii) Grounded generation depends on retrieval coverage and data freshness.
(iv) The current system is text-only and does not reason directly over EO and Earth Sciences imagery or structured geospatial data.

% ====================================================

\section*{CO\textsubscript{2} footprint}
\label{sec:co2}

% Synthetic Data Generation: This phase utilized 54,000 GPU hours, distributed across Marenostrum (19k hours), API services like OpenRouter(16k hours), and Mistral Compute (19k hours).

% Fine-tuning and Evaluation: The most compute-intensive phase was conducted on Mistral Compute using H100 80GB nodes, totaling 175,000 GPU hours.

% Training and operating large language models requires substantial computational resources, which in turn generate non-negligible carbon emissions. 

% This was distributed across public EuroHPC supercomputing clusters, API-based inference services, and Mistral’s dedicated compute infrastructure.
% We report the estimated carbon footprint associated with synthetic data generation, model fine-tuning, and evaluation stages of the EVE project.
% Aggregated across the various compute environments used, the total estimated carbon footprint of model development amounts to approximately 38 tonnes of $CO_2$ equivalent. This estimate accounts for energy consumption of GPU-based workloads and standard regional carbon intensity factors.
We estimate the carbon footprint of synthetic data generation, fine-tuning, and evaluation at about 38 tonnes of $CO_2$ equivalent, based on GPU energy consumption and regional carbon intensity factors.
% We emphasize that EVE is built on a compact 24B-parameter architecture, selected to balance domain performance with inference efficiency. This design choice enables cost-effective deployment and reduces long-term operational emissions compared to substantially larger general-purpose models.

% ====================================================

\section*{Acknowledgments}

This work was supported by ESA \(\Phi\)-lab under the Foresight Element of the FutureEO programme. We thank Imperative Space for their domain expertise, as well as Translated and Sapien for data annotation. We also thank Hiroshi Araki, Matteo Cacciola, Kumar Tulsi, and Eva Gmelich Meijling for their technical support during the development of EVE. We thank Andreas Vlachos for his guidance on hallucination detection.

\bibliography{custom}

% ====================================================
\appendix
\section{Corpus Creation}
\label{sec:appendix-corpus-creation}

We provide additional technical details, statistics, and validation results for the data collection and processing pipeline to build the corpus presented in Section~\ref{sec:data}. We assemble the corpus to cover the breadth of EO and Earth Sciences knowledge, including subtopics such as satellite imagery analysis, climate modeling, geospatial data processing, and environmental monitoring. 
The majority of sources are peer-reviewed domain-specific publishers (e.g., MDPI, NCBI), as well as reputable sources (e.g., ESA, NASA), and mainstream sources (Wikipedia, arXiv).
We present a full corpus distribution of the released open source data in Table~\ref{tab:data_distribution}.

\begin{table}[!htb]
    \small
    \centering
    \setlength{\tabcolsep}{4pt}
    \begin{tabular}{l r r l}
        \toprule
        Source & Released Tokens & \% & Licence \\
        \midrule
        MDPI               & 1.3 B   & 46.6   & CC-BY \\
        Copernicus         & 723 M   & 25.9   & CC-BY \\
        NCBI               & 485 M   & 17.4   & CC-BY \\
        ISPRS              & 74.1 M  & 2.7    & CC-BY \\
        Wiley              & 71.6 M  & 2.6    & CC-BY \\
        Elsevier           & 43.7 M  & 1.6    & CC-BY \\
        Cambridge Press    & 40.3 M  & 1.4    & CC-BY \\
        Springer           & 25.4 M  & 0.9    & CC-BY \\
        Taylor \& Francis  & 11.8 M  & 0.4    & CC-BY \\
        AMS                & 5.7 M   & 0.2    & CC-BY \\
        SAGE               & 3.8 M   & 0.1    & CC-BY \\
        NASA               & 2.5 M   & 0.09   & CC-BY \\
        arXiv              & 2.1 M   & 0.08   & CC-BY \\
        IEEE               & 992 k   & 0.04   & CC-BY \\
        EGUP               & 671 k   & 0.02   & CC-BY \\
        Oxford Academic    & 576 k   & 0.02   & CC-BY \\
        IOP Science        & 507 k   & 0.02   & CC-BY \\
        Frontiers          & 307 k   & 0.01   & CC-BY \\
        EOGE               & 245 k   & 0.01   & CC-BY \\
        EOS                & 96 k    & 0.003  & CC-BY \\
        MIT                & 83 k    & 0.003  & CC-BY \\
        UK Met Office      & 7 k     & 0.0002 & CC-BY \\
        \midrule
        Total              & \textbf{2.8 B} & \textbf{100} & \\
        \bottomrule
    \end{tabular}
    \caption{Token distribution across data sources in the released subset of the open-access EO and Earth Sciences corpus (2.8B tokens released out of 4.2B total open-access tokens).}
    \label{tab:data_distribution}
\end{table}

\textbf{Data Scraping.} We use Selenium Webdriver\footnote{\href{https://www.selenium.dev/documentation/webdriver/}{selenium.dev/documentation/webdriver/}} for automated browser navigation, paired with Brightdata Web Unlocker.\footnote{\href{https://brightdata.com/}{brightdata.com/}} This allows us to manage request rates, rotate IP addresses, solve captchas and maintain compliance with each websites. 

% Data Cleaning - 
% 1. Removal of any nougat artifacts - This component removes any artifacts or tags left by nougat ( eg: <WARNING>, <ERROR>)
% 2. Correction of merged words - This component handles any merged words where the word starts with a number appended to a word (eg: 1Introduction). These words are caught and a space is added between them. (1 Introduction)
% 3. OCR Duplication Removal – This component performs a MinHash based near duplicate search to identify repeated text segments. It then detects OCR introduced duplicates by checking for adjacent duplicates separated by minimal or no intervening characters, and removes them.
% 4. Rule-Based Removal – This component filters out lines composed of a single repeated symbol and normalizes excessive spacing by replacing three or more consecutive newline characters with two newlines.

\textbf{Data Cleaning.} We implemented a multi-stage data cleaning pipeline to improve corpus quality and remove extraction artifacts:

\begin{enumerate}[noitemsep, topsep=0pt]
    \item Nougat Artifact Removal \cite{blecher2023nougat}: we remove residual tags and artifacts introduced during PDF parsing (e.g., \texttt{<WARNING>}, \texttt{<ERROR>}).

    \item Merged Word Correction: we detect and correct tokenization errors where numeric prefixes are concatenated with words (e.g., \texttt{1Introduction} $\rightarrow$ \texttt{1 Introduction}).

    \item OCR Duplication Removal: we apply MinHash-based near-duplicate detection to identify repeated text segments. We further detect and remove OCR-induced duplicates via adjacency patterns (i.e., repeated spans with minimal or no intervening characters).

    \item Rule-Based Filtering: we remove low-information lines (e.g., sequences of repeated symbols) and normalize formatting by collapsing excessive whitespace (e.g., replacing three or more consecutive newlines with two).
\end{enumerate}

\textbf{Data Extraction.} To select an OCR system for scientific document extraction, we first construct a benchmark of 1k PDFs and evaluate multiple OCR tools. Ground-truth annotations are generated using a high-fidelity pipeline combining image encoding and GPT-4–based transcription. We measure OCR quality using the Normalized Levenshtein Similarity (NLS)~\cite{levenshtein1966binary}. 
Let $\hat{y}$ denote the predicted text and $y$ the ground-truth text. 
Let $\mathrm{LD}(\hat{y}, y)$ denote their Levenshtein distance, and $\mathrm{len}(\cdot)$ the sequence length. 
The metric is defined as:
\begin{equation}
\mathrm{NLS}(\hat{y}, y) = 
1 - \frac{\mathrm{LD}(\hat{y}, y)}
{\max\bigl(\mathrm{len}(\hat{y}), \mathrm{len}(y)\bigr)}.
\end{equation}

\noindent The NLS score ranges from 0 (no similarity) to 1 (exact match), quantifying agreement between OCR output and reference text. As shown in Table~\ref{tab:pdf_benchmarks}, several tools achieve high text-level similarity, but only Nougat consistently captures structured scientific content, including formulas and tables. This balance between textual fidelity and structural preservation, together with competitive latency (0.01s per page), motivates our selection for downstream processing.

\begin{table}[!htb]
    \centering
    \small
    \begin{tabular}{lcccc}
        \toprule
        Tool & Text & Formulas & Tables & Avg. \\
        \midrule
        Markitdown    & 0.81 & 0.00 & 0.00 & 0.27 \\
        Docling       & 0.79 & 0.00 & 0.40 & 0.40 \\
        Pymupdf4llm   & 0.81 & 0.00 & 0.26 & 0.36 \\
        Pypdf2        & \textbf{0.84} & 0.00 & 0.00 & 0.28 \\
        Marker        & 0.80 & 0.31 & \textbf{0.42} & 0.51 \\
        Unstructured  & 0.79 & 0.00 & 0.00 & 0.26 \\
        Pdfminer      & 0.81 & 0.00 & 0.00 & 0.27 \\
        Nougat        & 0.74 & \textbf{0.55} & 0.41 & \textbf{0.57} \\
        \bottomrule
    \end{tabular}
    \caption{OCR benchmark results (Normalized Levenshtein Similarity) across tools, evaluating text, formula, and table extraction. The Avg. column reports the mean across the three categories.}
    \label{tab:pdf_benchmarks}
\end{table}

% ====================================================

\section{Evaluation}
\label{sec:appendix-evaluation}

In addition to the category-level averages reported in Table~\ref{tab:gd-benchmarks} on general-domain benchmarks, we provide the full set of underlying benchmark results in Table~\ref{tab:eval_full}. These detailed results show that \texttt{EVE-Instruct} retains broad general capabilities after domain adaptation.

\begin{table}[!htb]
    \centering
    \small
    \setlength{\tabcolsep}{4pt}

    \begin{tabular}{@{}lrrr@{}}
        \toprule
        Benchmark & Small 3.2 & \texttt{EVE} & $\Delta$ \\
        \midrule

        aime25@16 & 26.7 & \textbf{35.2} & +8.5 \\
        aime24@16 & 37.1 & \textbf{41.2} & +4.1 \\
        math & \textbf{88.6} & 88.4 & -0.2 \\
        \textit{Average (Math \& Reasoning)} & \textit{50.8} & \textbf{\textit{54.9}} & \textit{+4.1} \\

        \midrule

        livecodebench\_p@1 & 36.4 & \textbf{39.1} & +2.7 \\
        mbpp\_p@1 & \textbf{74.7} & 73.9 & -0.8 \\
        \textit{Average (Coding)} & \textit{55.6} & \textbf{\textit{56.5}} & \textit{+0.9} \\

        \midrule
        
        super\_gpqa\_5shot & 38.8 & \textbf{40.5} & +1.7 \\
        mmlu\_redux\_5shot & 82.1 & \textbf{82.7} & +0.6 \\
        mmlu\_5shot & 80.6 & \textbf{81.7} & +1.1 \\
        mmlu\_astronomy\_5shot & 92.1 & \textbf{97.4} & +5.3 \\
        naturalqs\_5shot & \textbf{33.9} & 33.5 & -0.4 \\
        triviaqa\_5shot & \textbf{78.8} & 78.1 & -0.7 \\
        \textit{Average (Knowledge)} & \textit{67.7} & \textbf{\textit{69.0}} & \textit{+1.3} \\

        \midrule

        \textit{Average (Tool Calling)} & \textit{87.9} & \textbf{\textit{90.9}} & \textit{+3.0} \\
        
        \midrule
        
        \textit{Average (Instruction Following)} & \textit{80.1} & \textbf{\textit{81.2}} & \textit{+1.1} \\
        
        \midrule
        
        \textit{Average (Chat Quality)} & \textit{90.8} & \textbf{\textit{91.7}} & \textit{+0.9} \\

        \midrule

        \textit{Overall} & \textit{72.2} & \textbf{\textit{74.0}} & \textit{+1.8} \\

        \bottomrule
    \end{tabular}
    \caption{Evaluation results (0--100 scale) comparing Mistral Small 3.2 and \texttt{EVE-Instruct} across general-domain benchmarks. Category averages are shown for each task group. Tool Calling, Instruction Following, and Chat Quality correspond to private internal evaluation.}
    \label{tab:eval_full}
\end{table}

Additionally, we report results on the EO benchmarks in Table~\ref{tab:eve-tasks}, comparing \texttt{EVE-Instruct} with larger-scale models to complement the comparable-size comparisons in Table~\ref{tab:eo-benchmark-results-0-shot}. As shown in Table~\ref{tab:benchmark-results-0-shot-larger}, \texttt{EVE-Instruct} remains competitive even against substantially larger models, indicating strong efficiency in domain-specific performance.

\begin{table*}[htbp]
\centering
\small
\setlength{\tabcolsep}{3pt}
\begin{tabular}{@{}llccccccccc@{}}
\toprule

% & & \multicolumn{2}{c}{\scriptsize MCQA Multiple}
% & {\scriptsize MCQA Single}
% & {\scriptsize Hallucination}
% & \multicolumn{2}{c}{\scriptsize Open-Ended}
% & \multicolumn{2}{c}{\scriptsize Open-Ended w/ Context}
% & \\
% \cmidrule(lr){3-4}
% \cmidrule(lr){5-5}
% \cmidrule(lr){6-6}
% \cmidrule(lr){7-8}
% \cmidrule(lr){9-10}
% {\scriptsize Model} & {\scriptsize Size (B)} & {\scriptsize IoU} & {\scriptsize Acc.} & {\scriptsize Acc.} & {\scriptsize F1} & {\scriptsize Judge} & {\scriptsize EVE WR} & {\scriptsize Judge} & {\scriptsize EVE WR} & {\scriptsize Avg. Rank $\downarrow$} \\
% \midrule
 & & \multicolumn{2}{c}{\small MCQA Mult.} & {\small MCQA Sing.} & {\small Hallucin.} & \multicolumn{2}{c}{\small Open-Ended} & \multicolumn{2}{c}{\small Open-Ended w/ Context} & \\
\cmidrule(lr){3-4}
\cmidrule(lr){5-5}
\cmidrule(lr){6-6}
\cmidrule(lr){7-8}
\cmidrule(lr){9-10}
{\small Model} & {\small Size (B)} & {\small IoU} & {\small Acc.} & {\small Acc.} & {\small F1} & {\small Judge} & {\small EVE WR} & {\small Judge} & {\small EVE WR} & {\small Rank $\downarrow$} \\
\midrule
GPT-4.1 & 1800\textsuperscript{*} & \textbf{87.56} & 78.19 & 94.37 & 81.58 & 96.48 & 49.77 & \textbf{86.65} & 49.22 & 2.83 \\
Qwen3 & 235-A22 & 87.40 & \textbf{80.97} & 95.16 & 84.40 & \textbf{97.05} & 48.24 & 86.10 & 50.09 & \textbf{2.17} \\
MiniMax m2.5 & 230A10 & 84.82 & 77.72 & 94.95 & 83.77 & 91.00 & 51.10 & 81.57 & 51.20 & 5.17 \\
GPT OSS & 120A5 & 84.56 & 76.79 & 89.77 & 89.92 & 94.20 & 50.30 & 84.80 & 50.70 & 4.83 \\
Mistral Medium 3.1 & 200\textsuperscript{*} & 85.44 & 76.33 & 95.00 & 76.89 & 96.45 & 50.20 & 86.44 & 50.99 & 4.17 \\
GPT-5 nano & 20\textsuperscript{*} & 84.40 & 76.10 & 91.99 & \textbf{90.94} & 92.20 & 50.20 & 84.40 & 48.60 & 5.33 \\
\hdashline
EVE-Instruct & 24 & 86.12 & 77.73 & \textbf{96.35} & 84.70 & 96.40 & --- & 78.28 & --- & 3.50 \\
\bottomrule
\end{tabular}
\caption{Extension of Table~\ref{tab:eo-benchmark-results-0-shot} to larger-scale models under the same evaluation setup. Rank $\downarrow$ (lower is better) reports the average per-metric rank across MCQA multiple (IoU and Accuracy), MCQA single (Accuracy), Hallucination (F1), Open-ended QA (Judge), and Open-Ended QA with Context (Judge). \textsuperscript{*}Model size is reported when publicly available; otherwise estimated internally.}
\label{tab:benchmark-results-0-shot-larger}
\end{table*}

% \begin{table*}[htbp]
% \centering
% \small
% \setlength{\tabcolsep}{3pt}
% \begin{tabular}{@{}llccccccccc@{}}
% \toprule
%  & & \multicolumn{4}{c}{\small Open-Ended} & \multicolumn{4}{c}{\small Open-Ended w/ Context} & \\
% \cmidrule(lr){3-6}
% \cmidrule(lr){7-10}
%  & & \multicolumn{2}{c}{\small Claude Judge} & \multicolumn{2}{c}{\small Gemini Judge} & \multicolumn{2}{c}{\small Claude Judge} & \multicolumn{2}{c}{\small Gemini Judge} & \\
% \cmidrule(lr){3-4}
% \cmidrule(lr){5-6}
% \cmidrule(lr){7-8}
% \cmidrule(lr){9-10}
% {\small Model} & {\small Size (B)} & {\small Score} & {\small EVE WR} & {\small Score} & {\small EVE WR} & {\small Score} & {\small EVE WR} & {\small Score} & {\small EVE WR} & {\small Rank $\downarrow$} \\
% \midrule
% Llama Scout 4     & 109-A17 & 69.22 & 52.7 & 78.09 & 54.9 & 55.64 & 59.83 & 62.44 & 59.45 &  4.75 \\
% Qwen3             & 30-A3   & 82.75 & 51.2 & 88.89 & 51.2 & 69.13 & 53.71 & 64.59 & 53.23 &  1.75 \\
% Gemma 3           & 27      & 80.82 & 51.6 & 87.55 & 51.4 & 62.48 & 52.75 & 63.49 & 51.91 &  2.75 \\
% Mistral 3.2 Small & 24      & 73.76 & 51.0 & 84.04 & 52.5 & 57.21 & 57.19 & 60.28 & 56.69 &  4.25\\
% \hdashline
% EVE-Instruct      & 24      & 83.19 & ---  & 90.19 & ---  & 61.77 & ---   & 72.44 & ---   &  1.50 \\
% \bottomrule
% \end{tabular}
% \caption{Comparison of models on Open-Ended and Open-Ended with Context tasks using two judges (Claude and Gemini). Rank $\downarrow$ (lower is better) is the average per-metric rank across the four judge score columns.}
% \label{tab:benchmark-two-judge}
% \end{table*}

\begin{table*}[htbp]
\centering
\scriptsize
\setlength{\tabcolsep}{3pt}
\begin{tabular}{@{}llccccccccccc@{}}
\toprule
 & & \multicolumn{4}{c}{Open-Ended} & \multicolumn{4}{c}{Open-Ended w/ Context} & & & \\
\cmidrule(lr){3-6}
\cmidrule(lr){7-10}
 & & \multicolumn{2}{c}{Claude Judge} & \multicolumn{2}{c}{Gemini Judge} & \multicolumn{2}{c}{Claude Judge} & \multicolumn{2}{c}{Gemini Judge} & & & \\
\cmidrule(lr){3-4}
\cmidrule(lr){5-6}
\cmidrule(lr){7-8}
\cmidrule(lr){9-10}
{Model} & {Size (B)} & {Score} & {EVE WR} & {Score} & {EVE WR} & {Score} & {EVE WR} & {Score} & {EVE WR} & {Rank$_{\text{prev}}$ $\downarrow$} & {Rank $\downarrow$} & {$\Delta$Rank $\downarrow$} \\
\midrule
Llama Scout 4     & 109-A17 & 69.22 & 52.7 & 78.09 & 54.9 & 55.64 & 59.83 & 62.44 & 59.45 & 3.67 & 3.50 & $-$0.17 \\
Qwen3             & 30-A3   & 82.75 & 51.2 & 88.89 & 51.2 & 69.13 & 53.71 & 64.59 & 53.23 & 2.67 & 2.75 & $+$0.08 \\
Gemma 3           & 27      & 80.82 & 51.6 & 87.55 & 51.4 & 62.48 & 52.75 & 63.49 & 51.91 & 3.83 & 3.88 & $+$0.05 \\
Mistral 3.2 Small & 24      & 73.76 & 51.0 & 84.04 & 52.5 & 57.21 & 57.19 & 60.28 & 56.69 & 3.50 & 3.75 & $+$0.25 \\
\hdashline
EVE-Instruct      & 24      & 83.19 & ---  & 90.19 & ---  & 61.77 & ---   & 72.44 & ---   & \textbf{1.33} & \textbf{1.13} & $-$0.20 \\
\bottomrule
\end{tabular}
\caption{Comparison of models on Open-Ended and Open-Ended with Context tasks using two judges (Claude and Gemini - anthropic/claude-sonnet-4.6 and google/gemini-2.5-flash). Rank$_{\text{prev}}$ refers to the single-judge baseline ranking from Table~\ref{tab:eo-benchmark-results-0-shot}. Rank $\downarrow$ (lower is better) is the average per-metric rank across the four judge score columns. $\Delta$Rank denotes the difference between the two rankings.}
\label{tab:benchmark-two-judge}
\end{table*}

% \begin{table}[htbp]
% \centering
% \small
% \begin{tabular}{@{}lc@{}}
% \toprule
% {\small Model} & {\small Score T} \\
% \midrule
% Gemini 3.0 Flash    &  90.3\\
% Mistral 3.1 Small & 81.3\\
% \hdashline
% EVE-Instruct      & 82.3\\
% \bottomrule
% \end{tabular}
% \caption{Astrobench benchmark results.}
% \label{tab:score-t}
% \end{table}

Finally, we show that our domain adaptation and replay fine-tuning yield positive transfer to other technical domains, even without domain-specific training. We evaluate this in Table~\ref{tab:telqna-satcom-results}, in the telecommunications and satellite communications domain using both a multiple-choice QA benchmark, TelQNA \cite{maatouk2025teleqna}, and an open-ended QA dataset, Satcom Open-Ended \cite{corrente2026satcomllm}.

\begin{table}[!htb]
\centering
\small
\setlength{\tabcolsep}{3pt}
\begin{tabular}{@{}llcc@{}}
\toprule
 & & TelQNA & Satcom Open-Ended \\
\cmidrule(lr){3-3}
\cmidrule(lr){4-4}
Model & Size (B) & Acc. & Judge \\
\midrule
Llama4 Scout & 109-A17 & 74.57 & 75.00 \\
Qwen3 & 30-A3 & 73.81 & \textbf{83.00} \\
Gemma3 & 27 & 70.84 & 81.25 \\
Mistral Small 3.2 & 24 & 72.63 & 76.50 \\
\hdashline
\texttt{EVE-Instruct} & 24 & \textbf{75.72} & 81.25 \\
\bottomrule
\end{tabular}
\caption{Model performance on TelQNA and Satcom Open-Ended benchmarks (0-shot).}
\label{tab:telqna-satcom-results}
\end{table}

% ====================================================

\section{RAG Evaluation}
\label{sec:appendix-rag}

We detail the evaluation and design choices underlying the RAG pipeline introduced in Section~\ref{sec:grounded}. While the main text describes the system, we provide here a systematic analysis of chunking, embedding, and reranking to justify the final configuration.

We evaluate the impact of key design dimensions, including chunking strategy, embedding model, reranker, chunk size (512 vs.\ 1024), and quantization. In particular, we compare the hierarchical two-pass chunker (Section~\ref{sec:grounded}) against a standard fixed-length character chunker, and assess both quantized and non-quantized variants. For embedding and reranking, we focus on two representative models: Qwen3-Embedding-4B \cite{zhang2025qwen3embeddingadvancingtext}, a top-performing and efficient model on the MTEB leaderboard \cite{muennighoff2023mteb}\footnote{\href{https://huggingface.co/spaces/mteb/leaderboard}{huggingface.co/spaces/mteb/leaderboard}}, and INDUS \cite{bhattacharjee2024indus}, an encoder-only model trained specifically for scientific domains.

\subsection{Token-level Evaluation Framework}

Conventional information retrieval metrics emphasize document ranking order, yet large language models demonstrate relative insensitivity to where relevant content appears within their context window. Furthermore, when query-relevant information spans multiple chunks, inter-chunk ranking becomes inherently ambiguous. Drawing from Chroma's framework,\footnote{\href{https://research.trychroma.com/evaluating-chunking}{research.trychroma.com/evaluating-chunking}} we implement a token-granularity evaluation protocol for our retrieval pipeline.
We construct a semi-synthetic evaluation set by prompting an LLM to generate queries grounded in the corpus, along with corresponding relevant text excerpts. This approach avoids contamination of embedding models and enables domain-specific evaluation.
Each query–excerpt pair is evaluated using the following metrics:

% This semi-synthetic generation methodology offers several advantages. By creating evaluation data post-training, we eliminate potential contamination in general-purpose embedding models, thereby reducing evaluation bias. The approach also supports domain-adaptive assessment, enabling retrieval quality measurement across arbitrary datasets.

% Our generation process operates as follows: given our document corpus, we prompt an LLM to synthesize queries grounded in the corpus content while simultaneously identifying relevant text excerpts. 

% Adopting the Chroma framework, we evaluate retrieval performance through token-level metrics that align with the efficiency demands of LLM-powered systems. Each query-excerpt pair is assessed across multiple complementary quality dimensions.
% Our evaluation employs the following metrics:

\begin{itemize}[noitemsep,topsep=0pt]
    \item \textbf{Intersection over Union (IoU)}: Measures overlap between retrieved and ground-truth tokens. Penalizes redundancy when the same relevant tokens appear across multiple chunks.
    
    \item \textbf{Precision}: Token-level signal-to-noise ratio of retrieved tokens that are relevant. Reflects how much irrelevant context is introduced.
    
    \item \textbf{Recall}: Measures retrieval completeness by calculating the fraction of ground-truth relevant tokens successfully retrieved. Indicates whether the system captures all necessary information to answer the query.
    
    \item \textbf{Document Recall}: Percentage of documents containing at least one relevant chunk.
    
    \item \textbf{Passage Recall}: Fraction of retrieved chunks that contain at least one relevant token.
\end{itemize}

% \begin{itemize}[noitemsep,topsep=0pt]
    % \item \textbf{Intersection over Union (IoU)}: Primary efficiency measure quantifying the overlap ratio between ground-truth and retrieved tokens. Penalizes redundancy when overlapping chunking strategies cause identical relevant tokens to appear in multiple retrieved chunks, thereby inflating computational costs without information gain.
    
    % \item \textbf{Precision}: Token-level signal-to-noise ratio indicating the proportion of retrieved tokens that are actually relevant to the query. High precision minimizes irrelevant context retrieval, reducing both computational waste and potential LLM distraction.
    
    % \item \textbf{Recall}: Measures retrieval completeness by calculating the fraction of ground-truth relevant tokens successfully retrieved. Indicates whether the system captures all necessary information to answer the query.
    
    % \item \textbf{Document Recall}: Percentage of documents containing at least one relevant chunk, providing a coarse-grained view of retrieval coverage across the corpus.
    
    % \item \textbf{Passage Recall}: Proportion of retrieved chunks that contain at least one relevant token, measuring the hit rate at the chunk level rather than individual token granularity.
% \end{itemize}

Together, these metrics capture both retrieval quality and efficiency: token-level metrics (IoU, precision, recall) assess fine-grained relevance, while document and passage recall provide complementary coarse-grained coverage signals.

% These metrics collectively characterize both retrieval accuracy and computational efficiency. While IoU, precision, and recall operate at fine-grained token resolution, document retrieval ratio and passage recall provide complementary chunk-level and document-level perspectives on system performance. 

\subsection{Discussion}

\textbf{Embedder.} Table~\ref{tab:embedding_comparison_merged} compares embedding models across chunking strategies, sizes, and quantization. Quantization has negligible impact on retrieval quality, while providing clear gains in memory and inference efficiency.\footnote{We compute latency difference on a various kinds of retrievals on a subset, and observe between 66.6\% and 99.2\% reduction in latency in different setups when quantizing.}
Qwen3-Embedding-4B consistently outperforms INDUS embedder across all configurations, particularly in recall. Increasing chunk size to 1024 improves document and passage recall but degrades IoU and precision due to additional irrelevant tokens. Similarly, the Recursive chunker achieves higher recall but at the cost of substantially lower IoU, indicating increased redundancy.
We therefore select the Two-pass chunker with Qwen3-Embedding-4B. We fix a chunk size of 512, since qualitative evaluation by users in the chat interface consistently favored shorter, more focused chunks.

% \textbf{Reranker.} Table~\ref{tab:retrieval_reranking_top10} reports reranking results at top-10 with $K{=}20$ retrieved candidates. The Qwen3-4B reranker consistently improves over retrieval-only baselines across all embedding configurations, with the most pronounced gains in MRR (e.g., +6.7pp for Qwen3-Embedding-4B* at chunk size 512). In contrast, the INDUS reranker occasionally degrades performance when paired with stronger embeddings (Ref Retrieved Ratio drops from 63.30 to 62.80, MRR from 62.60 to 60.10), suggesting it is not well-calibrated for high-quality embedding spaces. The combination of 1024 chunk size with the Qwen3-4B reranker yields the best overall performance (MRR 73.40, Ref Retrieved Ratio 71.60). Despite this, we retain chunk size 512 in the final system: qualitative evaluation in the chat interface consistently favoured shorter, more focused chunks, and the MRR gap between the two configurations narrows considerably after reranking (69.30 vs.\ 73.40).

\textbf{Reranker.} Table~\ref{tab:retrieval_reranking_top10} reports reranking results at top-10 with $K{=}20$ retrieved candidates. Qwen3-4B-reranker consistently improves over retrieval-only baselines across all embedding configurations. In contrast, INDUS reranker can degrade performance when paired with stronger embeddings, suggesting weaker calibration in high-quality retrieval settings.
The best overall results are obtained with chunk size 1024 and the Qwen3-4B reranker. However, we retain chunk size 512 in the final system: qualitative evaluation favours shorter, more focused chunks, and the performance gap after reranking is relatively small.

\begin{table*}[h]
\small
\centering
\begin{tabular}{@{}lrlrrrrr@{}}
\hline
\textbf{Chunker} & \textbf{Chunk Size} & \textbf{Embedding} & \textbf{Doc recall} & \textbf{Passage recall} & \textbf{IoU} & \textbf{Precision} & \textbf{Recall} \\
\hline
Two-pass & 512 & INDUS 512 & 91.60 & 57.70 & 2.77 & 2.78 & 77.00 \\
\hline
Two-pass & 512 & INDUS 512* & 91.60 & 50.80 & 2.64 & 2.65 & 73.20 \\
\hline
Two-pass & 512 & INDUS 1024 & 83.24 & 38.60 & 2.34 & 2.35 & 59.90 \\
\hline
Two-pass & 512 & INDUS 1024* & 82.60 & 38.50 & 2.32 & 2.33 & 59.70 \\
\hline
Two-pass & 512 & Qwen3-Embedding-4B & 95.70 & 65.60 & 2.93 & \textbf{2.94} & 85.30 \\
\hline
Two-pass & 512 & Qwen3-Embedding-4B* & 95.60 & 63.60 & \textbf{2.94} & \textbf{2.94} & 85.00 \\
\hline
Two-pass & 1024 & INDUS 512* & 90.10 & 53.50 & 1.77 & 1.78 & 73.50 \\
\hline
Two-pass & 1024 & INDUS 1024* & 80.00 & 40.70 & 1.69 & 1.69 & 60.30 \\
\hline
Two-pass & 1024 & Qwen3-Embedding-4B* & \textbf{96.40} & 68.80 & 2.23 & 2.23 & 87.60 \\
\hline
Recursive & 1024 & INDUS 512* & 90.00 & 61.80 & 1.22 & 1.22 & 79.40 \\
\hline
Recursive & 1024 & INDUS 1024* & 78.50 & 44.60 & 1.04 & 1.05 & 63.70 \\
\hline
Recursive & 1024 & Qwen3-Embedding-4B* & 95.43 & \textbf{70.70} & 1.38 & 1.38 & \textbf{88.10} \\
\hline
\end{tabular}
\caption{Performance comparison of different embedding models across chunking strategies and chunk sizes. All metrics are expressed as percentages with two decimal precision. Precision and recall are per token, considering 10 as the number of passages retrieved. * indicates quantized embeddings. The Two-pass chunker refers to the approach presented in Section~\ref{sec:grounded}. The Recursive chunker is based on LangChain's \href{https://reference.langchain.com/python/langchain_text_splitters/\#langchain_text_splitters.RecursiveCharacterTextSplitter}{\texttt{RecursiveCharacterTextSplitter}}.}
\label{tab:embedding_comparison_merged}
\end{table*}

\begin{table*}[h]
\small
\centering
\begin{tabular}{@{}lrlrrrr@{}}
\hline
\textbf{Collection} & \textbf{Chunk Size} & \textbf{Reranker} & \multicolumn{2}{c}{\textbf{Ref Retrieved Ratio @10}} & \multicolumn{2}{c}{\textbf{MRR @10}} \\
\cline{4-7}
 &  &  & \textbf{Retrieval} & \textbf{Reranked} & \textbf{Retrieval} & \textbf{Reranked} \\
\hline
INDUS 512* & 512 & INDUS & 51.00 & 54.10 & 45.90 & 55.70 \\
\hline
INDUS 512* & 512 & Qwen3-4B & 51.00 & 54.70 & 45.90 & 61.70 \\
\hline
Qwen3-Embedding-4B* & 512 & INDUS & 63.30 & 62.80 & 62.60 & 60.10 \\
\hline
Qwen3-Embedding-4B* & 512 & Qwen3-4B & 63.30 & 65.10 & 62.60 & 69.30 \\
\hline
Qwen3-Embedding-4B* & 1024 & Qwen3-4B & \textbf{68.60} & \textbf{71.60} & \textbf{65.40} & \textbf{73.40} \\
\hline
Qwen3-Embedding-4B* & 1024 & INDUS & \textbf{68.60} & 65.90 & \textbf{65.40} & 61.00 \\
\hline
\end{tabular}
\caption{Performance comparison at top-10 with K=20 retrievals. 
\textbf{Ref Retrieved Ratio @10} measures the percentage of queries for which 
at least one relevant chunk appears in the top-10 reranked results. 
\textbf{MRR @10} (Mean Reciprocal Rank) is the average of the reciprocal rank 
of the first relevant chunk across queries, rewarding systems that place relevant 
chunks higher in the ranked list. All metrics are expressed as percentages with 
two decimal precision. * indicates quantized embeddings.}
\label{tab:retrieval_reranking_top10}
\end{table*}

% ====================================================

\section{System Architecture}
\label{sec:appendix-architecture}

EVE is deployed as a full-stack application comprising a React frontend, FastAPI backend, and a conversation management layer. 

\subsection{Conversation Management}

\textbf{Memory management}. To maintain conversational continuity, we use a rolling summarization strategy rather than retaining the full dialogue. At turn $t$ the model receives the previous turn $t-1$ in full, along with a compressed summary $S_{t_0}^{t-2}$ of all earlier turns. The most recent turn is always preserved verbatim to support immediate follow-up questions. Given the substantial length of each turn comprising the query, the generated answer, and the retrieved context a new summary is produced at every step by prompting \texttt{EVE-Instruct} to condense the previous summary and the latest turn: $S_{t_0}^{t-1} = \textit{summarize}(S_{t_0}^{t-2},\, t-1)$.

\begin{figure*}[!htb]
    \centering
    \includegraphics[width=1\linewidth]{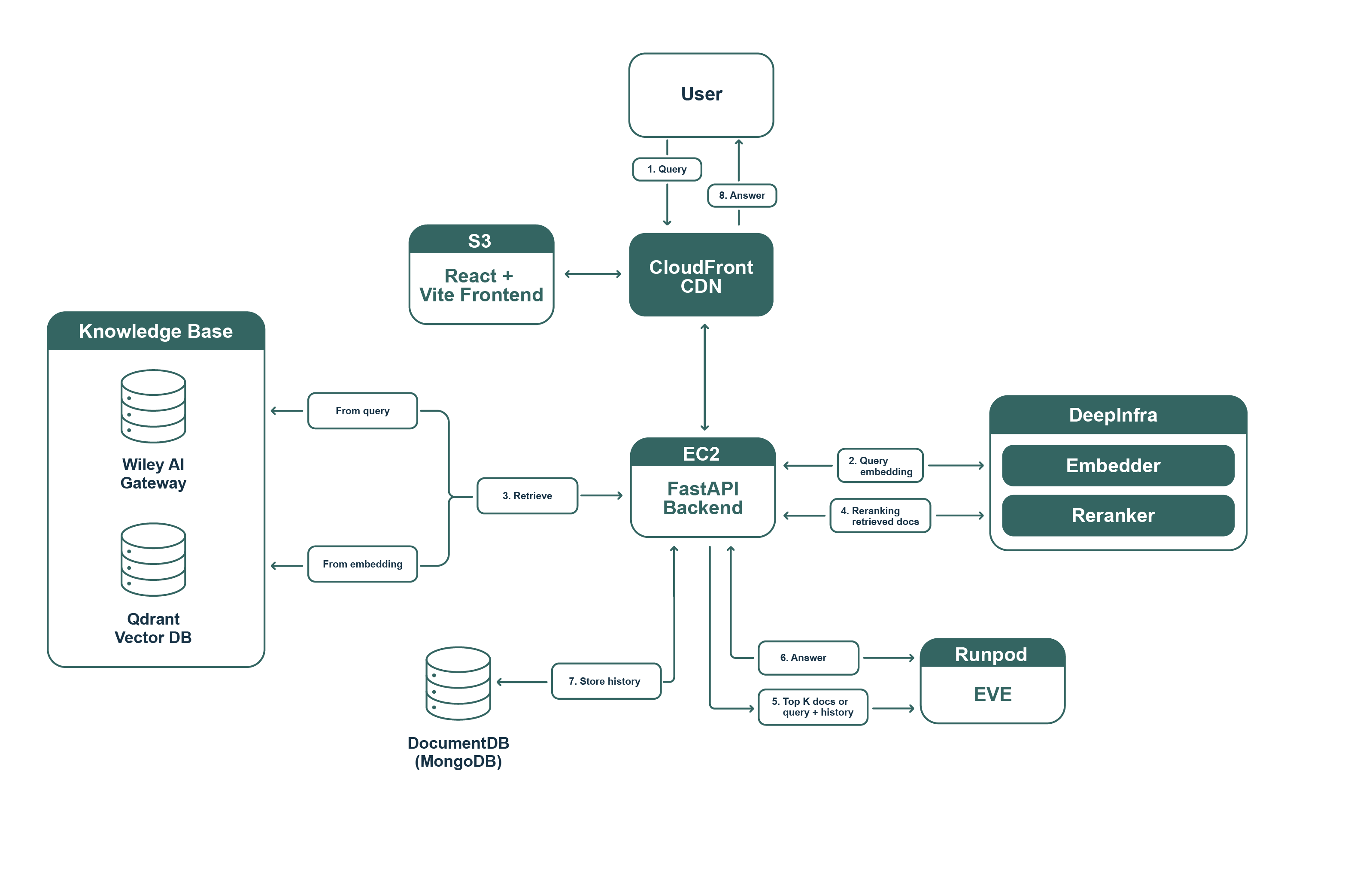}
    \caption{End-to-end architecture of the deployed EVE system.}
    \label{fig:userflow}
\end{figure*}

\textbf{Context Management.}  To balance the different components of the prompt, we enforce a fixed token budget:
% 
% (i) the user query is capped at 30K tokens and truncated if exceeded, (ii) the retrieved context is limited to 7K tokens, with low-similarity chunks dropped until the limit is met, (iii) the conversation summary is constrained to 5K tokens, enforced during summary generation, (iv) the model answer is allocated 15K tokens, and (v) the previous conversation turn is allocated 57K tokens.
% 
\begin{itemize}[noitemsep,topsep=0pt]
    \item \textbf{User query:} capped at 30K tokens and truncated if exceeded.
    \item \textbf{Retrieved context:} limited to 7K tokens, with low-similarity chunks dropped until the limit is met.
    \item \textbf{Conversation summary:} constrained to 5K tokens, enforced during summary generation.
    \item \textbf{Model response:} allocated 15K tokens.
    \item \textbf{Previous turn:} allocated 57K tokens.
\end{itemize}
Figure~\ref{fig:userflow} illustrates the end-to-end architecture of the deployed EVE system, including query processing, hybrid retrieval, re-ranking, grounded generation, and conversational state management.

\subsection{Backend}
\label{sec:backend}
The EVE backend is a FASTAPI\footnote{\href{https://github.com/fastapi}{github.com/fastapi}} service that handles data access, conversation state and document management. It is paired with Amazon DocumentDB for storage.

The state of each user is persisted with credentials, individual conversations and messages with timestamps, and the documents and collections used during retrieval. Authentication uses JWT tokens for protected routes along with CORS restriction. Every action performed on the interface is logged in a MongoDB instance. We have a dedicated internal dashboard that monitors user usage trends, feedback, types of queries and documents used, document level click rate and performance metrics.
We use uvicorn\footnote{\href{https://github.com/Kludex/uvicorn}{github.com/Kludex/uvicorn}} web server with multiple worker processes to handle concurrent requests. We also make use of lifespan hooks for database connections. The service is containerized using Docker.

% ====================================================

\subsection{Frontend}
\label{sec:frontend}

The EVE frontend is a single‑page React application built with TypeScript and Vite. The chat interface streams model responses as they are generated using Server‑Sent Events,\footnote{\href{https://developer.mozilla.org/en-US/docs/Web/API/Server-sent_events}{developer.mozilla.org/en-US/docs/Web/API/Server-sent\_events}} so users see answers appear token by token. Long conversations use list virtualization\footnote{\href{https://tanstack.com/virtual}{tanstack.com/virtual}} to stay fast even with many messages. A side panel shows the retrieved document chunks (with basic metadata) and lets users pin or remove sources. A settings view exposes key RAG and generation controls such as model choice, temperature, retrieval depth, and safety/tuning options.

The frontend manages server data with React Query\footnote{\href{https://tanstack.com/query}{tanstack.com/query}} and local UI state with React. Forms use React Hook Form\footnote{\href{https://react-hook-form.com}{react-hook-form.com}} and Zod\footnote{\href{https://zod.dev}{zod.dev}}  for client‑side validation. When a user sends a message, the client triggers retrieval and generation, then renders partial tokens as they arrive over SSE. Conversation metadata, settings presets, and recent sources are cached and refreshed on a schedule to balance freshness and responsiveness.

Errors from the backend are mapped to user‑friendly messages for common issues such as rate limits, context size limits, or empty retrieval results. For transient failures, the UI supports retries with backoff\footnote{\href{https://docs.aws.amazon.com/general/latest/gr/api-retries.html}{docs.aws.amazon.com/general/latest/gr/api-retries.html}} and allows users to cancel an in‑progress stream.

UI performance is improved with list virtualization, memoized content blocks, and deferring work for off‑screen panels. Vite provides code splitting and tree‑shaking to keep the bundle small. The production build is deployed via a GitHub Actions CI/CD pipeline to Amazon S3 and served through CloudFront with compression and aggressive caching for static assets. Runtime configuration is supplied via Vite environment variables, with secrets managed outside of version control.

% ====================================================

% ====================================================

% ====================================================

\section{Pilot Programme}
\label{sec:pilot}

The EVE platform underwent a structured pilot evaluation to assess its readiness as a domain-specific research assistant for the EO and Earth science community. This section describes the pilot setup and discusses the key findings that emerged from user engagement data and qualitative feedback.

% ---------------------------------------------------------------------------
% \subsection{Setup}\label{sec:pilot-setup}

The pilot programme enrolled 350 participants drawn from ESA technical staff, EO researchers, and affiliated stakeholders. Data collection combined multiple complementary methods: super-user interviews, one-to-one meetings with targeted stakeholders, structured questionnaires distributed to all participants, and direct usage telemetry from the EVE platform. The evaluation period was designed as an exploratory phase, encouraging users to test EVE across a broad range of EO queries rather than integrate it into daily workflows.

Participants spanned academia and research (37.7\%), industry and commercial (29.1\%), space and EO agencies (11.2\%), government (5.8\%), and students or unaffiliated users (13.5\%), with 78\% coming from research, industry, or agency backgrounds. In terms of EO expertise, 61.2\% identified as expert or very familiar, 21.3\% as moderately familiar, and 17.0\% as having low familiarity with the domain.

% During the pilot, users generated 450 conversations comprising 847 individual messages and uploaded 21 documents into EVE's knowledge base. The platform's RAG pipeline was active in 83.2\% of interactions, indicating that the vast majority of queries triggered retrieval from the underlying document collections. These collections included the Wiley AI Gateway, EVE's own open-access corpus, the ESA EO Knowledge Base, Wikipedia, and user-uploaded private collections.

Over the pilot period (September 2025 - March 2026), users generated 2,622 messages across 1,424 conversations, totalling approximately 6.8M input and 2.46M output tokens, with 40 documents uploaded. The RAG pipeline was active in 86.7\% of interactions. Monthly message volume peaked in September (943) and gradually declined through March (68), consistent with an exploratory evaluation. Knowledge base usage was distributed across the Wiley AI Gateway (42.7\%), EVE open-access corpus (40.3\%), ESA KB (7.3\%), Wikipedia EO (5.3\%), and user-uploaded documents (4.4\%). Document retrieval followed a heavy-tailed distribution: 60.7\% of documents were retrieved only once, while the most frequently accessed document was retrieved 380 times.
Mean latency was 15.62s for answer generation and 12.75s time to first token, with retrieval at 1.44s, re-ranking at 2.02s, and query embedding at 0.94s. System reliability, measured over a final-phase subset of 245 messages, showed 188/245 successful LLM calls, 211/245 successful retrievals, and 220/245 successful re-ranking operations.

Overall, the results position EVE as a high-potential domain-specific research assistant with clear relevance for EO applications, but with limited operational maturity at this stage. While the concept of domain-adapted LLMs for ESA scientific and engineering workflows was positively received, the pilot highlights key limitations in knowledge coverage, factual reliability, and human–AI interaction design. Addressing these aspects will be critical for enabling sustained deployment in professional settings.

\section{Compliance}
\label{sec:appendixx-compliance}

Beyond model architecture, EVE is designed as an open, European-aligned system, with efficiency and regulatory compliance treated as core design constraints. The compact architecture enables efficient deployment, and all components (model weights, data pipelines, and infrastructure) are released under open licenses where legally permissible.
In parallel, we conducted a structured compliance and governance analysis covering data sourcing, copyright, licensing, and responsible deployment. A detailed account is provided in a dedicated whitepaper, which presents an applied framework for developing and open-sourcing AI systems under regulatory constraints, including data provenance, anonymization, and retrieval-augmented architectures. We refer the reader to the full document for further details\footnote{\href{https://zenodo.org/records/18415713}{zenodo.org/records/18415713}}~\cite{hu2026responsible}.

% ====================================================

\section{Prompts}
\label{sec:appendix-prompts}

% For reproducibility, we also provide the prompt template used to evaluate Open-Ended with retrieval context under the LLM-as-a-judge framework (see Figure~\ref{fig:judge-prompt}). The judge model is conditioned on the question, the retrieved context (when applicable), and the candidate response, and assigns a score according to predefined evaluation criteria. Slight variations of this template are employed for pairwise Win Rate evaluation and for the no-context Open-Ended setting, where the retrieval passages are omitted. 
% We also provide in Figure~\ref{fig:gen-prompt}) the general prompt used for generation at inference.

For reproducibility, we provide the prompt templates used across evaluation and data generation pipelines. These prompts cover (i) evaluation under the LLM-as-a-judge framework (Section~\ref{sec:evaluation}) and (ii) synthetic data generation and filtering procedures (Section~\ref{sec:synthetic-data}). Minor variations of these templates are used across settings (e.g., with or without retrieval context, or for pairwise evaluation).

\subsection{LLM-as-a-Judge Evaluation Prompt}

For open-ended evaluation with retrieval context, we adopt an LLM-as-a-judge framework in which a judge model scores candidate responses conditioned on the question, reference answer, and retrieved context. The prompt used for this setting is shown in Figure~\ref{fig:judge-prompt}. Variants of this template are used for pairwise Win Rate evaluation and for no-context evaluation, where retrieval passages are omitted.

\begin{figure}[!htb]
\scriptsize
\begin{tcolorbox}[colback=white, colframe=black, title=\textbf{LLM-as-a-judge}, width=\linewidth, sharp corners, coltitle=black, colbacktitle=white]
\begin{verbatim}



You are evaluating answers about Earth Observation (EO) 
and Remote Sensing (RS).

**Domain Terminology Check**

Key abbreviations in EO/RS context:
- "EO" = Earth Observation - NOT "Essential Oils"
- "MSI" = Multispectral Instrument (e.g., Sentinel-2) -
NOT computer monitors
- "SAR" = Synthetic Aperture Radar -
NOT "Specific Absorption Rate"

**IMPORTANT: Accuracy Over Length**

Do NOT reward length or verbosity for its own sake.
A concise, accurate answer that captures the key facts
from the reference should
score EQUAL to a longer answer covering the same content.

- A 3-sentence accurate answer = A 10-sentence answer
with the same core information
- Only award higher scores for additional detail if it
adds meaningful, accurate
information beyond the reference
- Extra bullet points or formatting without additional
substance should NOT
increase the score

**Scoring Rules:**

- **Score 0-1**: Answer misinterprets core domain terms
OR contains major factual errors
- **Score 2**: Answer is vague or off-topic but doesn't
misinterpret terms
- **Score 3**: Answer is partially correct,
understands domain context
- **Score 4**: Answer correctly captures key facts
from the reference with good domain understanding
- **Score 5**: Answer is accurate, demonstrates clear
EO/RS domain expertise,
AND adds meaningful context

**Key Principle**: If the answer interprets domain
terms incorrectly
(e.g., "EO" as "Essential Oils", "MSI" as computer
monitors), score 0-1 regardless
of other content quality.

Question: "{question}"
Answer: "{output}"
Reference: "{reference}"

{format_instructions}
\end{verbatim}
\end{tcolorbox}
\caption{LLM-as-a-judge evaluation prompt for Open-Ended with retrieval context.}
\label{fig:judge-prompt}
\end{figure}

\subsection{Active Reading Chunk Filter}
\label{sec:appendix-ar-filter-prompt}

Prior to the Active Reading synthesis pipeline (Section~\ref{sec:synthetic-data}), corpus chunks are filtered by an LLM judge, in Figure~\ref{fig:ar-filter-prompt}. The judge assigns one of three ratings: \textit{Best} (high quality and highly relevant to EO), \textit{Good} (relevant but mediocre, or high quality but little related), or \textit{Bad} (poor quality or off-topic). Only \textit{Best}-rated chunks are passed to Active Reading; \textit{Good} chunks may appear in the raw long-form mixture.

\begin{figure}[!htb]
\scriptsize
\begin{tcolorbox}[colback=white, colframe=black, title=\textbf{Active Reading chunk filter prompt}, width=\linewidth, sharp corners, coltitle=black, colbacktitle=white]
\begin{verbatim}
You are an expert in Earth Observation and Remote Sensing.
Your task is to assess the quality and relevance of the
following text chunk for training a specialized LLM.

The model covers 25 EO sub-disciplines, including:
atmospheric science, earth monitoring, environmental
science, geospatial intelligence, LULC, LiDAR,
multispectral/hyperspectral imaging, SAR and InSAR,
hydrology, oceanography, cryosphere, agriculture,
forestry, disaster monitoring, urban planning, geology,
climate science, thermal sensing, photogrammetry,
GNSS and geodesy, and geoinformatics.

Assign one of three ratings:
- "Best": highly relevant to one or more sub-disciplines
  AND high quality (accurate, informative, well-written).
- "Good": relevant but mediocre quality, OR high quality
  but only tangentially related.
- "Bad": poor quality or unrelated to Earth Observation.

Text Chunk: {text}
\end{verbatim}
\end{tcolorbox}
\caption{Prompt used to filter corpus chunks before Active Reading synthesis. Only \textit{Best}-rated chunks enter the Active Reading pipeline.}
\label{fig:ar-filter-prompt}
\end{figure}

\subsection{Active Reading Strategy Generation}
\label{sec:appendix-ar-specific-prompt}

For task-specific Active Reading, the model is prompted to both generate questions from a source chunk and devise active learning strategies tailored to each question. The prompt used for this process is shown in Figure~\ref{fig:ar-specific-prompt}.

\begin{figure}[!htb]
\scriptsize
\begin{tcolorbox}[colback=white, colframe=black, title=\textbf{Active Reading strategy generation prompt}, width=\linewidth, sharp corners, coltitle=black, colbacktitle=white]
\begin{verbatim}
You are an expert in Earth Observation and Remote Sensing,
covering SAR, LiDAR, Multispectral Imaging, LULC,
Atmospheric Science, Hydrology, and more.

Your task has two parts:

1. Generate Questions: generate ~4 distinct, insightful
   questions answerable from the document, covering its
   key information and nuances.

2. Devise Strategies: for EACH question, devise 2 diverse
   active learning strategies that help deeply internalize
   the *type* of knowledge required, not the answer itself.

Example strategies:
- Conceptual Visualization: diagram a processing chain
  (e.g., DEM generation from an InSAR pair).
- Comparative Analysis: contrast C-band vs. L-band SAR
  for biomass estimation.
- Practical Scenario: plan which sensor to task for a
  flood response and justify the choice.
- Analogy: explain spectral signature using how the human
  eye distinguishes colors.
- Data Interpretation: outline a rule-based classifier
  for coastal land cover using NDVI and band ratios.
- Problem Formulation: write a research question with
  required data and expected outcome.

Be creative and go beyond these examples to maximize
deep, conceptual understanding.

Document: {text}
\end{verbatim}
\end{tcolorbox}
\caption{Prompt used for task-specific Active Reading. The model first generates questions from the source chunk, then devises two active learning strategies per question to guide synthesis.}
\label{fig:ar-specific-prompt}
\end{figure}

\subsection{Active Reading Predefined Strategy Selection}
\label{sec:appendix-ar-agnostic-prompt}

For predefined Active Reading, the model selects from a fixed set of predefined strategies based on strict eligibility rules applied to the source chunk. At most 2 strategies are selected per chunk. The selection prompt, including the rule-based criteria, is shown in Figure~\ref{fig:ar-agnostic-prompt}.

\begin{figure}[!htb]
\scriptsize
\begin{tcolorbox}[colback=white, colframe=black, title=\textbf{Active Reading predefined strategy selection prompt}, width=\linewidth, sharp corners, coltitle=black, colbacktitle=white]
\begin{verbatim}
You are an expert curriculum designer for an advanced
Earth Observation course. Your task is to be extremely
selective and choose only the most appropriate strategies
for the given document.

Document: {text}

STRICT SELECTION RULES — only select a strategy if the
document clearly and substantially meets its criteria:

1. paraphrastic_restatement: ONLY IF the document
   contains more than 100 words of dense technical
   information.
2. acronym_glossary: ONLY IF at least 6 technical
   acronyms are explicitly defined using the pattern
   Full Name (Acronym). Acronyms without this pattern
   or non-technical ones (e.g., USA, EU, AI, ML) do
   not count. Acronyms must relate to Earth Observation.
3. timeline_generation: ONLY IF the document contains
   at least 10 distinct dates, years, or time-related
   events.
4. workflow_description: ONLY IF the document explicitly
   describes a complex procedural sequence or steps.
5. technical_tutorial: ONLY IF the document's primary
   focus is to explain a specific, named technique
   in detail.

From the strategies that pass these rules, select at
most 2 of the most impactful ones.
\end{verbatim}
\end{tcolorbox}
\caption{Prompt used for predefined Active Reading. The model applies strict eligibility rules to select at most 2 strategies from a fixed predefined set.}
\label{fig:ar-agnostic-prompt}
\end{figure}

\subsection{Active Reading Data Generation}
\label{sec:appendix-ar-generation-prompt}

Once strategies are selected, each is applied to its source chunk to generate the final synthetic training document. The generation template is shown in Figure~\ref{fig:ar-generation-prompt}.

\begin{figure}[!htb]
\scriptsize
\begin{tcolorbox}[colback=white, colframe=black, title=\textbf{Active Reading data generation prompt}, width=\linewidth, sharp corners, coltitle=black, colbacktitle=white]
\begin{verbatim}
Here is a learning strategy:
{strategy}

Apply this strategy to the following document:
{text}

Generate only the resulting document based on the
strategy. Do not add any conversational text or
introductions.
\end{verbatim}
\end{tcolorbox}
\caption{Prompt used to generate synthetic training documents by applying a selected Active Reading strategy to a source chunk. The model is explicitly instructed to output only the resulting document.}
\label{fig:ar-generation-prompt}
\end{figure}

\subsection{SelfQA Generation}
\label{sec:appendix-selfqa-prompt}

SelfQA samples are derived from existing context-grounded QA pairs by reformulating them into fully self-contained questions that do not require access to a source document. The corresponding prompt is shown in Figure~\ref{fig:selfqa-prompt}.

\begin{figure}[!htb]
\scriptsize
\begin{tcolorbox}[colback=white, colframe=black, title=\textbf{SelfQA generation prompt}, width=\linewidth, sharp corners, coltitle=black, colbacktitle=white]
\begin{verbatim}
You are an expert in Earth Observation (EO). Your task
is to transform a context-grounded QA pair into a fully
self-contained sample suitable for instruction fine-tuning.

ORIGINAL PAIR:
- Question: {question}
- Answer: {answer}

Generate a new Question and Answer pair following these
rules strictly:

1. Self-Contained Question:
   - Must NOT require the source document to be understood.
   - Do NOT use phrases like "in the text" or "according
     to the document".
   - Integrate necessary context directly. For example,
     transform "What is its resolution?" into "What is
     the spatial resolution of the Sentinel-2 MSI sensor?"
   - Must be clear and unambiguous on its own.

2. High-Quality Answer:
   - Base the answer on the original answer and source
     document. Modify form, not content.
   - Must be correct, complete, and detailed.
   - Written in full, explanatory, pedagogical sentences.
\end{verbatim}
\end{tcolorbox}
\caption{Prompt used to generate SelfQA samples. Context-grounded QA pairs are reformulated into self-contained questions that can be answered from the model's parametric knowledge alone.}
\label{fig:selfqa-prompt}
\end{figure}

\subsection{ContextQA Quality Filtering}
\label{sec:appendix-contextqa-filter-prompt}

Generated ContextQA samples are filtered by an LLM judge following a two-step assessment: hard filters that immediately assign a \textit{Wrong} rating for critical failures, followed by quality evaluation. The five-point scale maps to the labels used in Table~\ref{tab:data_mix}: \textit{Best}, \textit{Good}, \textit{Mid}, \textit{Bad}, and \textit{Wrong}. The filtering prompt is shown in Figure~\ref{fig:contextqa-filter-prompt}.

\begin{figure}[!htb]
\scriptsize
\begin{tcolorbox}[colback=white, colframe=black, title=\textbf{ContextQA quality filter prompt}, width=\linewidth, sharp corners, coltitle=black, colbacktitle=white]
\begin{verbatim}
You are an expert quality analyst with deep knowledge in
Earth Observation (EO). Act as a strict quality gate for
a generated instruction-tuning sample.

INPUT:
- Question: {question}
- Answer: {answer}

ASSESSMENT FLOW:
1. Hard Filters: if the sample is not relevant to EO,
   has poor SFT style (not conversational or detailed),
   contradicts the source document, or is too trivial,
   assign Wrong immediately.
2. Quality Evaluation: if it passes, evaluate question
   quality and answer correctness and completeness.

RATING SCALE:
- Best (top ~1%): flawless. The question uncovers a deep
  or non-obvious aspect of the document. The answer is
  correct, complete, and exceptionally well-written with
  rich context, examples, or analogies.
- Good (top ~5%): strong but not Best. The question is
  non-trivial and well-posed. The answer is correct and
  complete but lacks the deeper insight of Best.
- Mid: usable, with minor flaws. The question may be
  slightly basic or the answer correct but too concise.
- Bad: significant issues. The answer is partially
  correct or incomplete, or the question is ambiguous.
- Wrong: fails a hard filter or is factually wrong.
\end{verbatim}
\end{tcolorbox}
\caption{Prompt used to filter ContextQA samples. A two-step assessment first applies hard filters, then evaluates quality on a five-point scale. \textit{Best} and \textit{Good} samples are retained for training as described in Table~\ref{tab:data_mix}.}
\label{fig:contextqa-filter-prompt}
\end{figure}

% \begin{figure*}[h]
% \begin{tcolorbox}[colback=white, colframe=black, title=\textbf{Chat prompt}, width=\textwidth, sharp corners, coltitle=black, colbacktitle=white]
% \begin{verbatim}
% You are a helpful assistant that helps researchers and students understand topics
% regarding Earth Observation. Answer the user's question based on the provided 
% context and any relevant information from the conversation history.

% Previous conversation: {conversation}

% Given the following context: {context}.

% Please reply in a precise and accurate manner to this query: {query}

% Answer:
% \end{verbatim}
% \end{tcolorbox}
% \caption{Generation prompt.}
% \label{fig:gen-prompt}
% \end{figure*}

\end{document}